\documentclass{article}

\usepackage[preprint]{icml2026}

\usepackage[T1]{fontenc}    %
\usepackage{hyperref}       %
\usepackage{booktabs}       %
\usepackage{amsmath}        %
\usepackage{amssymb}        %
\usepackage{amsfonts}       %
\usepackage{amsthm}         %
\usepackage{mathtools}      %
\usepackage{nicefrac}       %
\usepackage{microtype}      %
\usepackage{graphicx}
\usepackage{subcaption}
\usepackage{xcolor}

\graphicspath{{figures/}{./}}

\newcommand{\sw}[2]{\prescript{}{#1}{#2}}

\newcommand{\ie}{\mbox{\it i.e.}}

\newcommand{\SOthree}{\mathrm{SO}(3)}
\newcommand{\SOtwo}{\mathrm{SO}(2)}
\newcommand{\sphere}{S^2}
\newcommand{\Lsph}[1]{L^2_{#1}(\sphere)}       %
\newcommand{\Lrot}{L^2(\SOthree)}
\newcommand{\Mackey}[1]{\mathcal{M}_{#1}}      %

\newcommand{\lift}[1]{\Lambda_{#1}}            %
\newcommand{\proj}[1]{\Lambda_{#1}^{\dagger}}        %
\newcommand{\conv}{\star}                      %
\newcommand{\Cop}[2]{\mathcal{C}_{#1\to #2}}   %
\newcommand{\Rot}{\mathcal{R}}                 %
\newcommand{\piG}{\pi_{G}}
\newcommand{\pis}[1]{\pi_{#1}}

\newcommand{\dd}{\,\mathrm{d}}
\newcommand{\dmu}{\dd\mu}
\newcommand{\ii}{\mathrm{i}}
\DeclareMathOperator{\Id}{Id}

\newcommand{\liftsum}{\overline{\Lambda}}      %
\newcommand{\spinset}{\mathcal{S}}             %
\newcommand{\mraee}{\mathrm{MREE}}  %
\newcommand{\SSFNO}{\texorpdfstring{S\textsuperscript{2}FNO}{S\texttwosuperior FNO}}   %
\newcommand{\coderepo}{\url{https://github.com/astro-informatics/s2ai}}

\theoremstyle{plain}
\newtheorem{theorem}{Theorem}
\newtheorem{lemma}[theorem]{Lemma}
\newtheorem{proposition}[theorem]{Proposition}
\newtheorem{corollary}[theorem]{Corollary}
\theoremstyle{definition}
\newtheorem{definition}[theorem]{Definition}
\theoremstyle{remark}

\icmltitlerunning{Spin Spherical Fourier Neural Operators}

\begin{document}

\twocolumn[
  \icmltitle{Spin Spherical Fourier Neural Operators:\\Efficient Equivariant Learning for Vector and Spin Fields on the Sphere}

  \begin{icmlauthorlist}
    \icmlauthor{Kevin Mulder}{ucl}
    \icmlauthor{Matthew A.~Price}{ucl}
    \icmlauthor{Eirini Leftaki}{ucl}
    \icmlauthor{Jason D.~McEwen}{ucl,ati}
  \end{icmlauthorlist}

  \icmlaffiliation{ucl}{Department of Space \& Climate Physics, University College London, London, UK}
  \icmlaffiliation{ati}{The Alan Turing Institute, London, UK}

  \icmlcorrespondingauthor{Jason D.~McEwen}{jason.mcewen@ucl.ac.uk}

  \icmlkeywords{Spherical convolutions, Fourier neural operators, spin-weighted functions, equivariance, ICML}

  \vskip 0.3in
]

\printAffiliationsAndNotice{}

\begin{abstract}
  Many fields of physical interest on the sphere, such as wind, the polarisation of the cosmic microwave background and the shear of weak gravitational lensing, are spin fields, whose values change phase when the local frame rotates. We extend spherical Fourier neural operators (SFNOs) to fields of any spin, through lifted spherical convolutions that are exactly rotationally equivariant. Lifting spin fields to Mackey functions on the rotation group provides the conceptual framework, while computation is performed exactly and efficiently through their harmonic representation and by appealing to sampling theorems on the sphere and rotation group. We construct spin SFNOs (\SSFNO{}s) from these building blocks. We demonstrate empirically that rotational equivariance holds to machine precision for these layers. We achieve state-of-the-art performance on existing benchmarks, and introduce new benchmarks on which spin-aware models outperform their scalar counterparts.
\end{abstract}

\section{Introduction}
\label{sec:intro}

In many fields of science data are measured on the sphere, from the relic radiation of the Big Bang and the shapes of distant galaxies in cosmology, to the winds of the atmosphere and the currents of the ocean. Many of these data sets are not scalar. The horizontal wind and the ocean currents are vector, or spin-$1$, fields, while the polarisation of the cosmic microwave background (CMB) and the shear induced by weak gravitational lensing are spin-$2$ fields \citep{zaldarriaga1997all,castro2005weak,wallis2022mapping}. The value of a spin-$s$ field at a point is defined relative to a frame in the tangent plane and acquires the phase $e^{-\ii s\chi}$ when that frame rotates by $\chi$.

On the sphere, rotations play the role that translations play in the plane. Spherical CNNs consequently build on $\SOthree$-equivariant convolutions \citep{cohen2018spherical,esteves2018learning,kondor2018clebsch,cobb2021efficient,ocampo2023scalable}, and spherical Fourier neural operators (SFNOs) \citep{bonev2023spherical} bring the operator-learning perspective of Fourier neural operators \citep{li2021fourier,kovachki2023neural} to the sphere. When applied to vector or spin fields, however, scalar spherical CNNs and SFNOs typically consider the components of spin fields (\ie\ real and imaginary components) as unrelated scalar channels. This is the standard approach in data-driven weather models for wind \citep{pathak2022fourcastnet,bi2023accurate,lam2023learning,esteves2023scaling}, which does not respect general rotations.

Equivariant networks for spin fields exist \citep{esteves2020spin,xu2022unified,ballerin2025so3}, but at a price. Those that convolve on the full rotation group are costly in computation and parameters. The group CNN of \citet{ballerin2025so3}, for example, has about $2$ million parameters on spherical MNIST (whereas we achieve similar performance with $50$ thousand parameters). The equivariance of these networks, moreover, is asserted or illustrated rather than quantified, and benchmarks for spin fields are scarce.
Our layers build on \citet{mcewen2026unifying}, who makes the lift--convolve--project construction of \citet[Eq.~14]{cohen2019general} and \citet{xu2022unified} explicit on the sphere and shows that it unifies the spherical convolutions of the literature, so that our layers can learn filters of each of these forms.

Our contributions are fourfold. First, we extend spherical Fourier neural operators to fields of any spin (Sec.~\ref{sec:spinsfno}), building spin SFNOs (\SSFNO{}s) from two blocks, the spin spherical convolution and the convolution on the rotation group, so that rotational equivariance is exact for every convolutional layer and holds in the continuum for every activation. Second, we describe the layers by lifting spin fields to Mackey functions on the rotation group, which provides their conceptual framework and proofs \citep{mcewen2026unifying}, but compute them in harmonic space to ensure \SSFNO{}s are efficient in representation, computation and parameters (Sec.~\ref{sec:efficiency}). Third, we show that the convolutional layers of \SSFNO{}s are exactly equivariant on spin fields, with errors of order $10^{-15}$, that whole \SSFNO{} blocks and UNets with the activations and the normalisation removed are equivariant to round-off, and we discuss the equivariance errors introduced by the other layers (Sec.~\ref{sec:equivariance}). Fourth, we introduce three new public benchmarks for spin fields spanning classification, weather and cosmology (Sec.~\ref{sec:benchmarks}), on which our networks achieve state-of-the-art performance. Our layers, networks and benchmarks will be released with the \texttt{s2ai} code (\coderepo).

\section{Background and related work}
\label{sec:background}

\subsection{Spin fields and their harmonic analysis}
\label{sec:background:spin}

A spin-$s$ field $\sw{s}{f}$ is a complex function on the sphere whose values are defined relative to a local tangent frame and transform as
\begin{equation}
  \sw{s}{f}'(x) = e^{-\ii s\chi}\, \sw{s}{f}(x)
  \label{eq:spin-phase}
\end{equation}
under a right-handed rotation of the frame by $\chi$ \citep{newman1966note,goldberg1967spin}. We write $\sw{s}{f}_{\ell m} = \langle \sw{s}{f}, \sw{s}{Y}_{\ell m}\rangle$, with $\ell \ge \lvert s\rvert$, for its spin spherical harmonic coefficients and $F^{\ell}_{mn} = \langle F, D^{\ell *}_{mn}\rangle$ for the Wigner coefficients of $F \in \Lrot$, with the Wigner functions $D^{\ell}_{mn}$ in $zyz$ Euler angles $(\alpha,\beta,\gamma)$ and the unnormalised Haar measure. The coefficients of every spin rotate by the same Wigner matrices under the spin rotation operator $\Rot_\rho$ of \citet{mcewen2015directional} (Appendix~\ref{app:harmonic}). Fields are bandlimited at $L$ if their coefficients vanish for $\ell \ge L$, and functions on $\SOthree$ azimuthally bandlimited at $N$ if also $F^{\ell}_{mn} = 0$ for $\lvert n\rvert \ge N$. By appealing to the sampling theorems of \citet{mcewen2011novel} and \citet{mcewen2015novel}, the \texttt{s2fft} code \citep{price2024differentiable} transforms such signals exactly, at $\mathcal{O}(L^3)$ for spin spherical harmonic transforms and $\mathcal{O}(NL^3)$ for Wigner transforms.

\subsection{Neural operators and equivariant networks on the sphere}
\label{sec:background:operators}

The spherical Fourier neural operator (SFNO) of \citet{bonev2023spherical} learns one weight per spherical harmonic degree in the block structure of the Fourier neural operator (FNO) of \citet{li2021fourier}. Like the FNO, it has a continuous representation through harmonic space, since the harmonic coefficients of bandlimited signals, computed exactly by appealing to sampling theorems, give access to the underlying continuous operator, as in the continuous approaches to spherical convolution described by \citet{ocampo2023scalable}. Its layer is the non-localised zonal filter of \citet{esteves2018learning}, the isotropic scalar member of the generalised convolutions of \citet{cobb2021efficient}. Spherical CNNs that learn underlying continuous representations \citep{esteves2018learning,cobb2021efficient} are thus SFNOs, neural operators in the sense of \citet{kovachki2023neural}, although they did not carry the name (Appendix~\ref{app:lineage}). This point is also made for the discrete-continuous (DISCO) convolutions of \citet{ocampo2023scalable}, whose filters are continuous, in recent work that adopts them as local neural-operator layers \citep{liuschiaffini2024neural,bonev2025fourcastnet}.

On the rotation group, spherical CNNs apply pointwise non-linearities \citep{cohen2018spherical} or exact tensor products \citep{kondor2018clebsch}; \citet{cobb2021efficient} combine both and develop more efficient tensor-product activations. For spin fields, \citet{esteves2020spin} learn spin-changing harmonic filters, with non-linearities on each field, while \citet{xu2022unified} lift to the rotation group only for a pointwise non-linearity before a fixed projection, and \citet{ballerin2025so3} apply full group convolutions and pointwise non-linearities. In Secs.~\ref{sec:spinsfno} to~\ref{sec:benchmarks} we combine these ingredients in a neural operator for spin fields with the block structure of the SFNO, form the lift only where it is needed, for directional filters and the non-linear coupling of spins, measure equivariance on spin fields, and introduce benchmarks that include spin~$2$, whereas the experiments of the networks above use spins of at most~$1$.

\section{Equivariant spin convolutions by lifting}
\label{sec:foundation}

Our layers rest on the generalised lifted convolution of \citet{mcewen2026unifying}, which we recall concisely here, with the results that we use stated in Appendix~\ref{app:background}.

\subsection{Lifting spin fields to the rotation group}
\label{sec:lifting}

Spin-$s$ fields are the sections of the line bundle associated by the character $e^{\ii s\gamma}$ to the rotation group, the principal $\SOtwo$-bundle of frames over the sphere \citep{cohen2019general,weiler2021coordinate,mcewen2026unifying}, and correspond to the Mackey functions of type $s$ on the rotation group, those with $F(gR_z(\gamma))=e^{-\ii s\gamma}F(g)$. A spin-$s$ field is lifted to its Mackey function by $\lift{s}$ and returned by its adjoint, the projection $\proj{s}$, which integrates over the fibre against $e^{\ii s\gamma}$:
\begin{align}
  (\lift{s} f)(\alpha,\beta,\gamma) &= \frac{1}{\sqrt{2\pi}}\, e^{-\ii s\gamma}\, f(\beta,\alpha); \label{eq:lift-main}\\
  (\proj{s} F)(\beta,\alpha) &= \frac{1}{\sqrt{2\pi}} \int_0^{2\pi} e^{\ii s\gamma}\, F(\alpha,\beta,\gamma) \dd\gamma . \label{eq:proj-main}
\end{align} The lift is an isometry with $\proj{s}\lift{s} = \Id$, occupies the single Wigner column $n=-s$, $(\lift{s}f)^{\ell}_{mn} = (-1)^s\sqrt{8\pi^2/(2\ell+1)}\,\sw{s}{f}_{\ell m}\,\delta_{n,-s}$, and is equivariant, intertwining the regular representation $\pi_{\SOthree}$ with the spin rotation $\pis{s}(R_\rho) = \Rot_\rho$. Conversely, a function on $\SOthree$ is azimuthally bandlimited at $N$ if and only if it is a sum of at most $2N-1$ lifted spin fields, one for each $\lvert s\rvert<N$.

\subsection{Lifted convolutions in harmonic space}
\label{sec:convolutions}

We adopt the group convolution, which for $\Psi, F \in \Lrot$ reads:
\begin{align}
  (\Psi \conv F)(\rho) &= \int \Psi^{*}(\rho^{-1}\rho')\, F(\rho')\dmu(\rho') ; \label{eq:group-conv}\\
  (\Psi \conv F)^{\ell}_{mn} &= \sum\nolimits_{m'} F^{\ell}_{mm'}\, \Psi^{\ell *}_{nm'} . \label{eq:group-conv-harmonic}
\end{align}
The generalised lifted convolution lifts a spin field to the rotation group, convolves it there with a kernel $\Psi$ and projects the result back to a spin field on the sphere. It resolves to four cases depending on whether a lift is applied to the input and a projection to the output:
\begin{align}
  \Cop{s}{s'}[\Psi] &= \proj{s'}(\Psi \conv \lift{s}\,\cdot\,), \notag\\
  \Cop{s}{\SOthree}[\Psi] &= \Psi \conv \lift{s}\,\cdot\,, \notag\\
  \Cop{\SOthree}{s'}[\Psi] &= \proj{s'}(\Psi \conv \cdot\,), \notag\\
  \Cop{\SOthree}{\SOthree}[\Psi] &= \Psi \conv \cdot\,.
  \label{eq:spin-conv-def}
\end{align}
All four are equivariant for every square-integrable kernel, and by the single-column lift only one kernel entry per degree reaches a spin-$s'$ output \citep{mcewen2026unifying},
\begin{equation}
  \big(\Cop{s}{s'}[\Psi]\, \sw{s}{f}\big)_{\ell m} = (-1)^{s+s'}\; \sw{s}{f}_{\ell m}\; \Psi^{\ell *}_{-s',-s} ,
  \label{eq:spin-multiplier}
\end{equation}
so that the spin spherical convolution $\Cop{s}{s'}$ maps spin $s$ to spin $s'$ by one multiplier per degree, without forming a signal on the rotation group. Specialising the equivariant kernel theorems \citep{kondor2018generalization,kondor2018clebsch,cohen2019general,xu2022unified} to the sphere shows that these multipliers are complete, in the sense that every bounded equivariant complex-linear map $T$ from spin-$s$ to spin-$s'$ fields acts as
\begin{equation}
  (T\,\sw{s}{f})_{\ell m} = c_\ell\;\sw{s}{f}_{\ell m}
  \label{eq:completeness}
\end{equation}
for a unique bounded sequence $(c_\ell)_{\ell\ge\max(\lvert s\rvert,\lvert s'\rvert)}$, and that every such sequence defines one \citep{mcewen2026unifying}.

A directional layer $\Cop{s}{\SOthree}[\Psi]$ whose kernel column $m'=-s$ is supported on $\lvert n\rvert<N$ is exactly a tuple of at most $2N-1$ spin-changing layers, which relates the directional layers of \citet{cobb2021efficient} to the multi-spin layers of \citet{esteves2020spin}.

A key result of \citet[Corollary~V.7]{mcewen2026unifying} is that directional kernels, a spherical output of one spin and rotational equivariance cannot be achieved simultaneously. Between two fixed spins the kernel of an equivariant map is radial apart from the frame phases, so that a layer whose output is a field of one spin cannot be both directional and equivariant. Directional information must therefore be kept on the rotation group or, equivalently, in a tuple of spin fields.

Since the generalised lifted convolution reduces to the spherical convolutions of the literature, each recovered by the support of its kernel, the choice of spins and whether the lift or the projection is omitted \citep{mcewen2026unifying}, layers built on it can learn filters of each of these forms rather than fix one by design. The spin spherical convolution contains every bounded equivariant complex-linear map between two spins, among them the axisymmetric convolutions of scalar spherical CNNs, the left convolution of \citet{driscoll1994computing}, the spin-weighted convolutions of \citet{esteves2020spin} and the spherical Kaiser--Squires relation of weak lensing \citep{kaiser1993mapping,wallis2022mapping}, and the convolution on the rotation group contains the directional, steerable and group convolutions \citep{wiaux2006fast,mcewen2015directional,cohen2018spherical,cobb2021efficient}. In our networks the filters are restricted by their parameterisation, through interpolated profiles in $\ell$ and, on the rotation group, a small internal spin set and a kernel diagonal on it.

\section{Spin spherical Fourier neural operators}
\label{sec:spinsfno}

\begin{figure*}[t]
  \centering
  \includegraphics{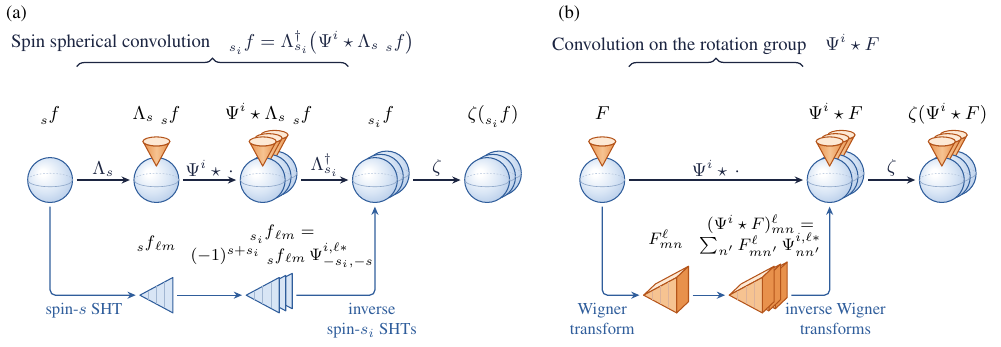}
  \caption{The spin spherical convolution (a), one multiplier per degree between spin spherical harmonic transforms (SHTs) without forming the lift, and the convolution on the rotation group (b), each followed by $\zeta$ and computed in harmonic space below. Spheres denote fields on $\sphere$, spheres with a cone denote functions lifted to $\SOthree$, triangles denote spin spherical harmonic coefficients and prisms Wigner coefficients, and $i$ indexes output spins $s_i$ in (a) and kernels in (b).}
  \label{fig:building-blocks}
\end{figure*}

We build \SSFNO{}s from two blocks with exactly equivariant linear layers, joined by the lift $\lift{s}$ and its adjoint $\proj{s}$.

\subsection{Convolutional building blocks}
\label{sec:blocks}

The spin spherical convolution (Fig.~\ref{fig:building-blocks}a) maps a tuple of spin fields to another by one multiplier \eqref{eq:spin-multiplier} per degree for each pair of spins and channels, without explicitly forming the lift. It provides complete linear spin mixing (Sec.~\ref{sec:convolutions}) but no non-linear coupling of spins, since its non-linearity acts on each field separately.

The convolution on the rotation group (Fig.~\ref{fig:building-blocks}b) acts on the Wigner coefficients of a function on $\SOthree$ by \eqref{eq:group-conv-harmonic}, and its non-linearity acts pointwise on samples on the group, with $2N-1$ samples in $\gamma$. In our networks its kernel is diagonal on an internal spin set $\spinset$, so that the linear part filters each spin separately and the spins are coupled by the non-linearity on the group, which is much less constrained than one on the sphere (Sec.~\ref{sec:nonlinearity}). The projection $\proj{s}$ then reads the Wigner column $n=-s$ of each spin $s\in\spinset$, and our implementation computes only these columns, at $\mathcal{O}(\lvert\spinset\rvert(L^3+NL^2))$ rather than the $\mathcal{O}(NL^3)$ of full Wigner transforms (Sec.~\ref{sec:computation}).

\subsection{Equivariant non-linearities}
\label{sec:nonlinearity}

Activations for spin fields may be applied in two settings, to the spin fields themselves on the sphere or pointwise to their lift on the rotation group; each setting brings its own constraint for equivariance. On the sphere a rotation turns the local frame and multiplies a spin-$s$ value by the phase of \eqref{eq:spin-phase}, so that for $s\neq0$ a pointwise activation is equivariant only if it commutes with that phase, $\zeta(e^{\ii\phi}z)=e^{\ii\phi}\zeta(z)$, and may therefore act on the modulus alone, as magnitude, norm and gated non-linearities do \citep{worrall2017harmonic,weiler2018three,esteves2020spin}. On $\SOthree$ rotations act on the argument alone, so that every pointwise map is equivariant \citep{cohen2018spherical,cobb2021efficient,ballerin2025so3}, with a much weaker constraint. A tuple of spin fields may therefore be lifted and summed, passed through any pointwise map and returned (Appendix~\ref{app:nonlinearity}).

On the sphere, for non-zero spin fields, we adopt the modulus non-linearity
\begin{equation}
  \zeta(z) = \mathrm{ReLU}\big(a\lvert z\rvert + b\big)\,\frac{z}{\lvert z\rvert+\epsilon},
  \label{eq:modrelu}
\end{equation}
with a learned real gain $a$ and offset $b$, and $\epsilon=10^{-8}$, which keeps the phase and is therefore equivariant for every spin. For scalar fields on the sphere ($s=0$) we adopt standard activation functions. On $\SOthree$ we apply a pointwise activation function to samples of the lifted tuple, such as the modulus non-linearity, and return to the sphere by the fixed projection $\proj{s}$ followed by a learned filter for each spin.

\subsection{\SSFNO{} blocks}
\label{sec:spinsfno-block}

\begin{figure}[t]
  \centering
  \includegraphics{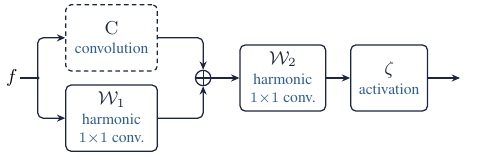}
  \caption{The \SSFNO{} block \eqref{eq:spinsfno-block}, with $\mathrm{C}$ the spin spherical convolution or the convolution on the rotation group, $\mathcal{W}_1$ and $\mathcal{W}_2$ harmonic $1\times1$ convolutions and $\zeta$ the activation (in detail in Fig.~\ref{fig:spinsfno-block-detail}).}
  \label{fig:spinsfno-block}
\end{figure}

The \SSFNO{} block (Fig.~\ref{fig:spinsfno-block}), acting on a tuple $f$ of spin fields,
\begin{equation}
  f \mapsto \zeta\big(\mathcal{W}_2(\mathrm{C}\,f+\mathcal{W}_1f)\big),
  \label{eq:spinsfno-block}
\end{equation}
with $\mathrm{C}$ the convolution of either building block, follows the SFNO block of \citet{bonev2023spherical} in summing a harmonic-space convolution and a pointwise branch, but replaces the MLP that follows the sum by $\mathcal{W}_2$ and has no identity residual connection. The operators $\mathcal{W}_1$, in parallel with the convolution, and $\mathcal{W}_2$, after the sum, are harmonic $1\times1$ convolutions. Each transforms its input to harmonic space, truncates or pads the coefficients to the output bandlimit, mixes the channels with one matrix $W$ and the spins with one matrix $M$, both constant in $\ell$, and transforms back; on $\SOthree$ it reads the columns $n=-s$, mixes them and lifts again. Between equal spins this is exactly the pointwise $1\times1$ convolution of the bypass of the FNO. Between different spins it keeps the harmonic coefficients and changes their spin, which in pixel space is a normalised power of the spin raising or lowering operator \citep{mcewen2026unifying}; since this is not local, it is computed in harmonic space. The role of the harmonic $1\times1$ convolutions is to enrich the channel and spin mixing of the block.

\subsection{Networks}
\label{sec:architecture}

We compare two constructions, each with Conv blocks, a convolution and its non-linearity, or with \SSFNO{} blocks, and name the four models by block and construction, Conv-$\sphere$, Conv-$\SOthree$, \SSFNO{}-$\sphere$ and \SSFNO{}-$\SOthree$. The spherical construction never lifts explicitly, whereas the rotation-group construction lifts after a first spin spherical convolution, so that its filters act on the rotation group. We construct classifier and dense prediction networks based on these architectural building blocks. The classifiers stack six blocks and pool rotation-invariant features, and we construct UNets with encoders and decoders built from the same blocks, with skip connections that concatenate features at equal bandlimit and spin set (Fig.~\ref{fig:unet}). The blocks include a normalisation, a group normalisation in the UNets and a batch normalisation in the classifiers, and the spin UNets for wind and weak lensing are Conv-$\SOthree$ networks (Appendix~\ref{app:implementation}).

\section{Efficient representation, computation and parameterisation}
\label{sec:efficiency}

Mackey functions and the lift give the conceptual description of the layers and their proofs \citep{mcewen2026unifying}, whereas the harmonic representation provides exact and efficient computation. We form the lift explicitly only where it provides expressivity.

\subsection{Representation and spin mixing}
\label{sec:representation}

In harmonic space the lift need not be formed. The lift of a spin-$s$ field occupies a single Wigner column and so is held in the $\mathcal{O}(L^2)$ spin spherical harmonic coefficients of the field rather than the $\mathcal{O}(L^3)$ of a general function on $\SOthree$, and a tuple of fields with spins in $\spinset$ in $\mathcal{O}(\lvert\spinset\rvert L^2)$ coefficients. Only an explicit lift, its synthesis at $2N-1$ samples in $\gamma$ for a pointwise non-linearity, requires $\mathcal{O}(NL^2)$.

Linear spin mixing needs no lift. Every bounded equivariant complex-linear map between two spins is a spin spherical convolution, and a tuple of spin fields, one for each spin $\lvert s\rvert<N$, is exactly a function on $\SOthree$ of azimuthal bandlimit $N$ \citep{mcewen2015novel}. A group convolution mapping between such functions is therefore an array of $(2N-1)^2$ spin spherical convolutions, one for each pair of spins, so that, as classes of linear maps, the spherical construction realises every convolution on $\SOthree$ at that azimuthal bandlimit without forming the lift. The convolution on the rotation group in our networks uses only the diagonal of this array, so that its spins are coupled by the non-linearity alone, and the rotation-group construction mixes spins linearly in its first layer. A pointwise complex-linear map cannot mix spins equivariantly, since under a frame rotation \eqref{eq:spin-phase} fields of different spins acquire different phases, so that an equivariant $1\times1$ convolution on the samples mixes channels only within a spin and the pointwise bypass of the FNO, extended to spin fields, must mix spins in harmonic space.

General non-linear spin coupling is what the explicit lift achieves. On the sphere it needs maps whose outputs have a definite spin, such as the modulus $z\mapsto\lvert z\rvert$, which takes every spin to spin $0$ and is the phase collapse of spherical scattering networks \citep{mcewen2022scattering,mousset2024generative}, or products of fields, whose spins add. The modulus non-linearity \eqref{eq:modrelu} keeps the phase, and hence the spin, of each field, so that the non-linearities of our spherical construction couple no spins. On $\SOthree$ every pointwise map is equivariant, and one non-linearity after the lift couples all spins at once. In the rotation-group construction spins are thus coupled non-linearly through the lift.

Representing a vector or spin field as a spin field, rather than as scalar channels, also matters under reflections of the sphere.\label{sec:parity} Networks on scalar fields built from isotropic multipliers and pointwise maps, without positional inputs, are $\mathrm{O}(3)$-equivariant \citep{weiler2021coordinate} and cannot represent maps that anticommute with the antipodal map $P$, such as the advection of a vorticity treated as a scalar. Spin fields escape this restriction, since $P$ reverses the orientation of the tangent frame and so acts on them through conjugation as well as pull-back (Appendix~\ref{app:parity}). A multiplier between spins commutes with $P$ if and only if its coefficients are real, so that with real profiles our classifiers would be exactly invariant under reflections, and they therefore use complex profiles. Real profiles, whose multipliers commute with $P$, could instead be useful for fields with antipodal symmetry, such as those of diffusion magnetic resonance imaging \citep{descoteaux2007regularized}.

\subsection{Computation}
\label{sec:computation}

We leverage sampling theorems and fast harmonic transforms on $\sphere$ and $\SOthree$ \citep{mcewen2011novel,mcewen2015novel,price2024differentiable}. Since the sampling theorems make the transforms of bandlimited spin fields and functions on $\SOthree$ exact, every linear layer acting on bandlimited input is computed exactly, consistent with the machine-precision equivariance errors of Sec.~\ref{sec:equivariance}. A spin spherical convolution costs $\mathcal{O}(L^3)$ for each field, and so $\mathcal{O}(\lvert\spinset\rvert L^3)$ for a tuple, against $\mathcal{O}(NL^3)$ for a convolution on $\SOthree$ at azimuthal bandlimit $N$, whose output fills all $2N-1$ spins, and $\mathcal{O}(L^4)$ for a full convolution on $\SOthree$, with $N=L$. These are upper bounds for the algorithms in harmonic space \citep{mcewen2026unifying}. Against convolutions that fill every spin, the saving of the spherical construction lies in a spin set smaller than $2N-1$ and in storing no samples on the rotation group, which the rotation-group construction samples only for its non-linearity. This non-linearity, which couples all spins, costs $\mathcal{O}(NL^3)$ with full Wigner transforms and $\mathcal{O}(\lvert\spinset\rvert(L^3+NL^2))$ in our implementation, which computes only the Wigner columns that are read.

\subsection{Parameterisation}
\label{sec:parameters}

We parameterise every linear layer in harmonic space. Our multipliers are interpolated in $\ell$ from ten knots placed at fixed fractions of the bandlimit, so that, as with the anchor points of \citet{cobb2021efficient}, filters at high resolution need few parameters. The knots reduce the $\mathcal{O}(L)$ multipliers of a spin spherical convolution for each pair of spins and channels to ten, against the $\mathcal{O}(NL)$ values of a kernel column of azimuthal bandlimit $N$ and the $\mathcal{O}(L^3)$ of a full kernel on $\SOthree$ \citep{mcewen2026unifying}. In the UNets the profiles are real, with a channel matrix for each output spin, and in the classifiers complex (Appendix~\ref{app:implementation}). Each harmonic $1\times1$ convolution has one channel matrix and one spin matrix, and the activations learn an offset for each channel and, in the UNets, a gain as well. Parameters therefore attach, through the knots, to fractions of the bandlimit rather than to samples, so that their number depends neither on the bandlimit nor on the sampling grid, and a trained layer applies on any grid that supports exact transforms at its bandlimit. Our layers are thus neural operators in the sense of \citet{kovachki2023neural}.

Our layers are built on the \texttt{s2fft} code \citep{price2024differentiable} and will be released with the \texttt{s2ai} code.

\section{Equivariance}
\label{sec:equivariance}

The convolutional layers of \SSFNO{}s are exactly equivariant to rotations of spin fields, which we confirm numerically, and we discuss the equivariance errors introduced by the other layers, such as the activations and the normalisation. Equivariance errors have been measured before for spherical CNNs on scalar signals \citep{cohen2018spherical,esteves2018learning,cobb2021efficient,ocampo2023scalable}, but not for vector and spin fields on the sphere under general rotations.

\subsection{Sources of equivariance error}
\label{sec:error-sources}

As for spherical CNNs on scalar signals, convolutional layers are computed exactly in harmonic space. Any error therefore enters only through the non-linear pointwise activations and the normalisation. The output $\zeta(F)$ of an activation is not bandlimited, so that analysing its samples on the same grid aliases its content above the bandlimit, and on $\SOthree$ beyond the azimuthal bandlimit. Oversampling reduces this aliasing \citep{cobb2021efficient}. The group normalisation of the UNets aliases for every spin, because it rescales the modulus by $\lvert x-\mu\rvert/\lvert x\rvert$, up to constants for each group (Appendix~\ref{app:normalisation}), and on the sphere it adds, for non-zero spin, a structural error, since the mean that it subtracts is not rotation invariant, of the kind that pointwise spin mixing would cause (Sec.~\ref{sec:representation}). A normalisation with statistics computed in harmonic space and no mean subtracted for $s\neq0$ would be exactly equivariant, which we leave to future work.

\subsection{Equivariance tests}
\label{sec:experiments:equivariance}

We measure the mean relative equivariance error ($\mraee$), a quadrature-weighted relative $L^2$ error, over $100$ random bandlimited signals and $100$ random rotations at $L=16$ in double precision (Appendix~\ref{app:equivariance-error}). Linear layers attain errors of order $10^{-15}$, the activations alias with errors of order $0.1$ to $1$, apart from the complex ReLU, measured at initialisation where it is the identity up to $\epsilon$, and the group normalisation gives errors of order $10^{-2}$ on the sphere and $10^{-3}$ on the rotation group (Table~\ref{tab:equivariance}). With the activations and the normalisation removed, whole \SSFNO{} blocks and UNets, with spin and channel mixing, are equivariant to the round-off floor of the transforms, of order $10^{-14}$ at the larger bandlimit $L=32$ (Tables~\ref{tab:equivariance} and~\ref{tab:equiv-linear}), which confirms that the errors of our blocks and networks arise only from their activations and normalisation.

\begin{table}[t]
  \caption{Equivariance errors in double precision on the sphere ($\sphere$) and the rotation group ($\SOthree$), measured by the mean relative equivariance error ($\mraee$) between rotating the input and rotating the output (Appendix~\ref{app:equivariance-error}). Conv-$\sphere$ and Conv-$\SOthree$ are the spin spherical convolution $\Cop{s}{s'}$ and the directional convolution $\Cop{s}{\SOthree}$, ReLU and Swish are applied directly to complex data, whereas the complex activations act on the modulus with the phase kept \eqref{eq:modrelu}, and the complex ReLU is measured at initialisation, where it is the identity up to $\epsilon$. The \SSFNO{} blocks and the UNets of the spin spherical MNIST architecture have no activation or normalisation, are at $L=32$, with $N=8$ on $\SOthree$, and reach the round-off floor of the transforms. All other rows are at $L=16$, with $N=5$ on $\SOthree$.}
  \label{tab:equivariance}
  \centering\footnotesize
  \setlength{\tabcolsep}{2.5pt}
  \begin{tabular}{llcc}
    \toprule
    & & \multicolumn{2}{c}{Equivariance error} \\
    \cmidrule(lr){3-4}
    Operation & Spin & $\sphere$ & $\SOthree$ \\
    \midrule
    Conv-$\sphere$ & $0$ & $5.9\times10^{-15}$ & -- \\ %
    Conv-$\sphere$ & $1$ & $5.2\times10^{-15}$ & -- \\ %
    Conv-$\SOthree$ & $2$ & -- & $6.0\times10^{-15}$ \\ %
    Harmonic $1\times1$ conv. & & $5.4\times10^{-15}$ & $5.4\times10^{-15}$ \\ %
    \SSFNO{} block & $\{0,\pm1\}$ & $2.1\times10^{-14}$ & $2.4\times10^{-14}$ \\
    \SSFNO{} UNet & $0{\to}1$ & $2.0\times10^{-14}$ & $1.8\times10^{-14}$ \\
    \SSFNO{} UNet & $1{\to}0$ & $2.0\times10^{-14}$ & $1.6\times10^{-14}$ \\
    \midrule
    ReLU (complex data) & & $0.453$ & $0.592$ \\ %
    Swish (complex data) & & $0.581$ & $1.047$ \\ %
    Complex ReLU, init. & & -- & $2.2\times10^{-9}$ \\ %
    Complex Swish & & -- & $0.0885$ \\ %
    Group norm. & & $0.013$ & $0.0033$ \\ %
    \bottomrule
  \end{tabular}
\end{table}

\section{Spin benchmarks and experiments}
\label{sec:benchmarks}\label{sec:experiments}

In this section we first introduce three new benchmarks for spin fields, spin spherical MNIST, global wind forecasting and weak gravitational lensing, and then compare our networks on each of them with scalar counterparts and published results (Appendix~\ref{app:training}).

\subsection{New benchmarks}
\label{sec:datasets}

Public benchmarks for learning on spin fields are lacking. The benchmarks of spherical CNNs, such as spherical MNIST \citep{cohen2018spherical}, are scalar, and the vector-field MNIST of \citet{esteves2020spin}, built from the Sobel gradients of the images and also used by \citet{xu2022unified}, is limited to spin~$1$. We introduce three new benchmarks, a contribution of this article in their own right, whose inputs and targets have a definite spin, of any spin for spin spherical MNIST, and which span classification and regression on synthetic, atmospheric and cosmological fields. Their generation code will be released with \texttt{s2ai}, and Appendix~\ref{app:datasets} describes them in full.

\paragraph{Spin spherical MNIST.} MNIST digits \citep{lecun1998gradient} are projected onto the sphere \citep{cohen2018spherical} at $L=32$. The Sobel-gradient construction of \citet{esteves2020spin}, the only one available so far, gives a spin-$1$ field only. We instead synthesise spin fields of any spin from the same harmonic coefficients in the spin-$s$ basis, so that rotations act exactly on the bandlimited fields, and use the harmonic spin-$1$ and spin-$2$ fields alongside the Sobel spin-$1$ field. We consider the classification of the digit from one of these fields, and dense prediction, which maps the spin-$0$ image to its spin-$1$ field or back and so changes spin. Both follow the NR/R protocol of training on unrotated and testing on randomly rotated data \citep{cohen2018spherical,cobb2021efficient,ocampo2023scalable}, which tests the invariance of a classifier and the equivariance of a dense prediction to rotations not seen in training (Appendix~\ref{app:mnist}).

\paragraph{Global wind forecasting.} The task is to predict the horizontal wind at 50~hPa at 18:00~UTC on a given day from the wind at 00:00, 06:00 and 12:00~UTC of the same day, with the wind represented by the spin-$1$ field $u+\ii v$ of its zonal and meridional components at $L=128$. The data are the ERA5 reanalysis \citep{hersbach2020era5} of WeatherBench~2 \citep{rasp2024weatherbench2} from 1980 to 2022, from which eight days drawn at random are held out in every year, so that the benchmark tests interpolation within the same years rather than generalisation to later years (Appendix~\ref{app:data:wind}).

\paragraph{Weak gravitational lensing.} The task is to reconstruct the spin-$0$ convergence $\kappa$ on the full sky from the noisy spin-$2$ shear $\gamma=\gamma_1+\ii\gamma_2$. Matter along the line of sight, most of it dark matter, distorts the images of distant galaxies by the shear, which on the full sky is a spin $0\to2$ multiplier of the convergence, a projection of the matter density \citep{bartelmann2001weak,kilbinger2015cosmology}. The spherical Kaiser--Squires estimator inverts this multiplier as a spin $2\to0$ spin spherical convolution \citep{kaiser1993mapping,wallis2022mapping}, but practical considerations make the recovery of $\kappa$ ill-posed and call for priors learned from simulations, which on the full sky have so far taken the two shear components as scalar channels \citep{wang2026advancing}. We simulate $10{,}000$ pairs of convergence and noisy shear maps for training and validation and $600$ for evaluation at $L=256$ with GLASS \citep{tessore2023glass} and the shape noise of a Stage~IV survey \citep{laureijs2011euclid} (Appendix~\ref{app:massmapping}).

\begin{table}[t]
  \caption{Spin spherical MNIST classification: NR/R accuracy (\%) of our spherical ($\sphere$) and rotation-group ($\SOthree$) networks (best accuracy and fewest parameters in bold).}
  \label{tab:mnist-classification}
  \centering\footnotesize
  \setlength{\tabcolsep}{3pt}
  \begin{tabular}{lrccc}
    \toprule
    & & Sobel & \multicolumn{2}{c}{Harmonic} \\
    \cmidrule(lr){3-3}\cmidrule(lr){4-5}
    Model & Params & spin 1 & spin 1 & spin 2 \\
    \midrule
    Conv-$\sphere$              & \textbf{50k} & 99.15 & 99.19 & 99.17 \\
    \SSFNO{}-$\sphere$          & 182k & 98.82 & 99.16 & \textbf{99.34} \\
    Conv-$\SOthree$             & 134k & 99.01 & 99.10 & 99.11 \\
    \SSFNO{}-$\SOthree$         & 136k & 99.17 & \textbf{99.21} & 99.28 \\
    \midrule
    \citeauthor{esteves2020spin}        & 58k  & 98.2  & -- & -- \\
    \citeauthor{ballerin2025so3}, sCNN  & 2.2M & 99.20 & -- & -- \\
    \citeauthor{ballerin2025so3}, GsCNN & 2.0M & \textbf{99.32} & -- & -- \\
    \bottomrule
  \end{tabular}
\end{table}

\subsection{Spin spherical MNIST}
\label{sec:experiments:mnist}

We train four models as classifiers and as UNets for dense prediction (Appendix~\ref{app:implementation}). The classifiers are selected on a non-rotated validation split and the prediction UNets on the rotated held-out set, which is also their evaluation set. The Conv-$\sphere$ classifier on the harmonic spin-$1$ field is trained with three seeds, whose mean accuracy we report, and the architecture of the classifiers was developed with reference to their test accuracies (Appendix~\ref{app:training:mnist}).

The four classifiers generalise to rotations not seen in training (Table~\ref{tab:mnist-classification}). On the Sobel field our best classifier comes within $0.15\%$ of the published accuracy of the general spherical CNN (GsCNN) of \citet{ballerin2025so3}, and on the harmonic spin-$1$ and spin-$2$ fields, for which no other results are published, all four exceed $99\%$. The classifiers use complex harmonic filters, since with real profiles they would be exactly invariant under reflections of the sphere (Sec.~\ref{sec:parity}).

\begin{figure}[t]
  \centering
  \includegraphics[width=\columnwidth]{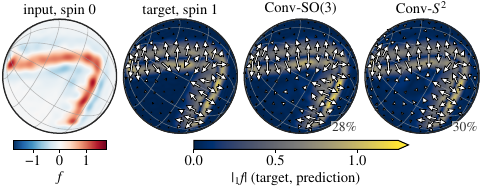}
  \caption{Prediction of the spin-$1$ Sobel field from the spin-$0$ image of a rotated test digit, chosen in advance, by the Conv-$\SOthree$ and Conv-$\sphere$ UNets, with their relative errors. Both predictions follow the rotated frame of the target (more in Fig.~\ref{fig:mnist-pred-full}).}
  \label{fig:mnist-pred}
\end{figure}

In dense prediction, a spin-changing task, each rotation-group UNet has fewer parameters and a lower error than its spherical counterpart in all four mappings (Table~\ref{tab:mnist-prediction} and Fig.~\ref{fig:mnist-pred}). Moreover, the Conv-$\SOthree$ UNet gives the lowest errors of the models compared from images to fields, and the \SSFNO{}-$\SOthree$ UNet from fields to images. The published values, for another construction and given for context only, lie above those of the Conv-$\SOthree$ UNet, which has about eight times fewer parameters than that of \citet{esteves2020spin}.

\begin{table}[t]
  \caption{Spin-changing prediction on spin spherical MNIST: NR/R MSE from the spin-$0$ image to its spin-$1$ field ($0\to1$) and back ($1\to0$), with the best model and the fewest parameters in bold.}
  \label{tab:mnist-prediction}
  \centering\footnotesize
  \setlength{\tabcolsep}{3pt}
  \begin{tabular}{lrcccc}
    \toprule
    & & \multicolumn{4}{c}{MSE $\times10^{3}$ $\downarrow$} \\
    \cmidrule(lr){3-6}
    & & \multicolumn{2}{c}{Sobel field} & \multicolumn{2}{c}{Harmonic field} \\
    \cmidrule(lr){3-4}\cmidrule(lr){5-6}
    Model & Params & $0\to1$ & $1\to0$ & $0\to1$ & $1\to0$ \\
    \midrule
    Conv-$\sphere$         & 23k & 3.681 & 3.212 & 5.580 & 3.229 \\
    \SSFNO{}-$\sphere$     & 29k & 4.253 & 4.544 & 3.944 & 4.468 \\
    Conv-$\SOthree$        & \textbf{14k} & \textbf{3.016} & 3.144 & \textbf{2.876} & 3.101 \\
    \SSFNO{}-$\SOthree$    & 20k & 3.866 & \textbf{2.948} & 3.063 & \textbf{3.005} \\
    \midrule
    \citeauthor{esteves2020spin} & ${\approx}112$k & 4.3 & 4.0 & -- & -- \\
    \citeauthor{xu2022unified}   & --              & 3.8 & --  & -- & -- \\
    \bottomrule
  \end{tabular}
\end{table}

\subsection{Global wind forecasting}
\label{sec:experiments:wind}

\begin{table}[t]
  \caption{Wind forecasting at 50~hPa on the held-out days and mass mapping from the noisy spin-$2$ shear of weak lensing at $L=256$, averaged over 600 maps (best results and fewest parameters in bold; all metrics in Tables~\ref{tab:wind-full} and~\ref{tab:massmapping-full}). All models are spherical, and the scalar UNet takes the spin field as two scalar channels. Persistence forecasts the 12:00~UTC field for 18:00~UTC.}
  \label{tab:wind}
  \label{tab:massmapping}
  \centering\footnotesize
  \setlength{\tabcolsep}{3pt}
  \begin{tabular}{lrcccc}
    \toprule
    Model & Params & MSE $\downarrow$ & SNR $\uparrow$ & PSNR $\uparrow$ & PCC $\uparrow$ \\
          &        &                  & (dB) & (dB) & \\
    \midrule
    \multicolumn{6}{l}{\emph{Global wind forecasting} (MSE in m$^2$\,s$^{-2}$)} \\
    Spin UNet        & \textbf{48k} & \textbf{9.52} & \textbf{14.02} & \textbf{25.06} & \textbf{0.979} \\
    Scalar UNet      & 74k & 17.27 & 11.43 & 22.64 & 0.969 \\
    Persistence      & --  & 14.55 & 12.17 & 23.19 & 0.962 \\
    \midrule
    \multicolumn{6}{l}{\emph{Weak gravitational lensing} (MSE $\times10^{6}$)} \\
    Spin UNet        & \textbf{81k} & \textbf{1.28} & \textbf{7.53} & \textbf{23.47} & \textbf{0.907} \\
    Scalar UNet      & 93k & 2.06 & 5.45 & 21.40 & 0.846 \\
    \bottomrule
  \end{tabular}
\end{table}

\begin{figure}[!t]
  \centering
  \includegraphics[width=\columnwidth]{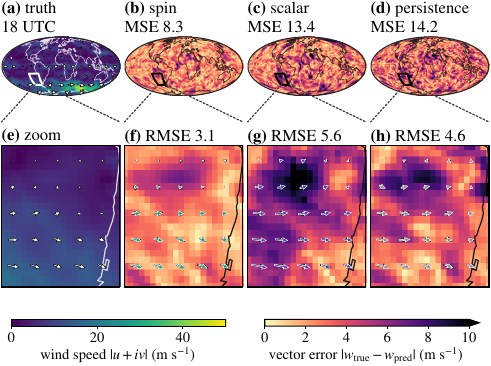}
  \caption{Wind forecast for one held-out day: (a) the target speed, (b--d) the vector errors of the spin and scalar UNets and persistence, and (e--h) the same in the outlined region, with true (white) and predicted (coloured) vectors. The scalar UNet leaves larger, coherent patches of error (more in Fig.~\ref{fig:wind-full}).}
  \label{fig:wind}
  \vspace{\floatsep}

  \centering
  \includegraphics[width=\columnwidth]{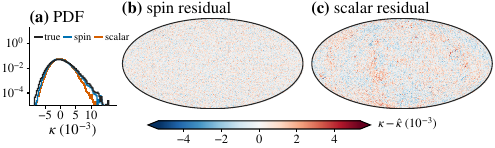}
  \caption{Mass mapping for one evaluation map at $L=256$: (a) PDFs of the true and reconstructed convergence, and residuals $\kappa-\hat\kappa$ of (b) the spin and (c) the scalar UNet. The scalar UNet misses the tails and leaves large-scale structure in its residual, whereas the spin residual is noise-like (maps in Fig.~\ref{fig:massmap-full}).}
  \label{fig:massmap}
\end{figure}

We compare a spin UNet, a Conv-$\SOthree$ network that takes the spin field as input, with a scalar spherical UNet that takes its real and imaginary parts as scalar channels, and with persistence, which forecasts the 12:00~UTC field for 18:00~UTC (Appendix~\ref{app:architectures}). Checkpoints are selected on the held-out days, which are also the evaluation set. The spin UNet reduces the MSE by $44.9\%$ with $34.9\%$ fewer parameters and improves every other metric (Table~\ref{tab:wind} and Fig.~\ref{fig:wind}). It is the only model that improves on persistence in every metric, whereas the scalar UNet is worse than persistence in every metric except the correlation. These results support the practical value of the spin representation.

\subsection{Weak gravitational lensing}
\label{sec:experiments:massmapping}

Table~\ref{tab:massmapping} and Fig.~\ref{fig:massmap} compare a spin UNet, which lifts the spin-$2$ shear and returns the spin-$0$ convergence, with a scalar UNet of DISCO convolutions \citep{ocampo2023scalable} and residual blocks \citep{he2016deep} that takes $\gamma_1$ and $\gamma_2$ as scalar channels, both five-level UNets \citep{ronneberger2015unet} (Appendix~\ref{app:architectures}). The spin UNet reduces the MSE by $37.9\%$ with $13.4\%$ fewer parameters and improves every other metric.

\section{Conclusions}
\label{sec:conclusion}

We have developed \SSFNO{}s (spin spherical Fourier neural operators), which extend spherical Fourier neural operators to fields of any spin, with rotational equivariance exact for every convolutional layer and holding in the continuum for every activation. Described by Mackey functions but computed in harmonic space, \SSFNO{}s are efficient in representation, computation and parameters, with far fewer parameters than networks that convolve spin fields on the full rotation group. On spin fields their convolutional layers are equivariant to $10^{-15}$, and whole blocks and UNets without activations or normalisation to round-off. On three new benchmarks our networks achieve state-of-the-art results.
\label{lastmainpage}%

\clearpage
\section*{Impact Statement}
This paper presents work whose goal is to advance the field of machine learning, and in particular the modelling of geophysical and astrophysical fields on the sphere. Improved treatment of spin fields such as wind and polarisation may benefit weather, climate and cosmological applications. There are many potential societal consequences of our work, none of which we feel must be specifically highlighted here.

\section*{Declaration on the use of large language models}
LLMs were used in the writing of this manuscript, but the theory was worked out manually. We take full responsibility for the accuracy of the material presented.

\bibliography{refs}
\bibliographystyle{icml2026}

\newpage
\appendix
\onecolumn

\section{Background: harmonic analysis, lifting and lifted convolutions}
\label{app:background}

In this appendix we collect the conventions and results of \citet{mcewen2026unifying} on which our layers rest, each cited in its heading by its number there, where it is proved. Definition~\ref{def:lift}, Lemma~\ref{lem:lift}, Corollary~\ref{cor:spin-tuple}, Theorem~\ref{thm:harmonic} and Proposition~\ref{prop:directional-tuple} are stated in the same words, and Theorem~\ref{thm:completeness} and Corollary~\ref{cor:nogo} in shortened form. Only the equivariance of the lifted non-linearity is proved in this article (Appendix~\ref{app:nonlinearity}).

\subsection{Notation and conventions}
\label{app:harmonic}

We adopt the conventions of \citet[Sec.~II and App.~A]{mcewen2026unifying}. Points of the sphere $\sphere$ are written $\omega=(\theta,\varphi)$, with colatitude $\theta\in[0,\pi]$ and longitude $\varphi\in[0,2\pi)$, and $\eta=(0,0)$ denotes the north pole. We adopt the measure $\dmu(\omega)=\sin\theta\dd\theta\dd\varphi$ and the inner product $\langle f,g\rangle=\int_{\sphere}f\,g^{*}\dmu$. Rotations are parameterised by $zyz$ Euler angles $\rho=(\alpha,\beta,\gamma)$ in the active convention, $R(\alpha,\beta,\gamma)=R_z(\alpha)R_y(\beta)R_z(\gamma)$, with the unnormalised Haar measure $\dmu(\rho)=\sin\beta\dd\alpha\dd\beta\dd\gamma$ of total mass $8\pi^2$. The map $p(\rho)=R_\rho\eta$ sends $R(\alpha,\beta,\gamma)$ to the point $(\beta,\alpha)$, so that $\sphere\cong\SOthree/\SOtwo$, with $\SOtwo=\{R_z(\gamma)\}$ the stabiliser of the north pole.

The spherical harmonics $Y_{\ell m}$ carry the Condon--Shortley phase, $Y^{*}_{\ell m}=(-1)^mY_{\ell,-m}$, and the spin raising and lowering operators $\eth$ and $\bar\eth$ \citep{newman1966note,goldberg1967spin} generate the spin spherical harmonics from them by
\begin{equation}
  \eth\,\sw{s}{Y}_{\ell m} = \sqrt{(\ell-s)(\ell+s+1)}\;\sw{s+1}{Y}_{\ell m},
  \qquad
  \bar\eth\,\sw{s}{Y}_{\ell m} = -\sqrt{(\ell+s)(\ell-s+1)}\;\sw{s-1}{Y}_{\ell m},
  \qquad \sw{0}{Y}_{\ell m} = Y_{\ell m},
  \label{eq:eth}
\end{equation}
so that $\sw{s}{Y}_{\ell m}$ exists only for $\ell\ge\lvert s\rvert$ and, for fixed $s$, these form an orthonormal basis of the space $\Lsph{s}$ of square integrable spin-$s$ functions on the sphere.

\begin{figure}[htbp]
  \centering
  \includegraphics[width=0.5\textwidth,page=2]{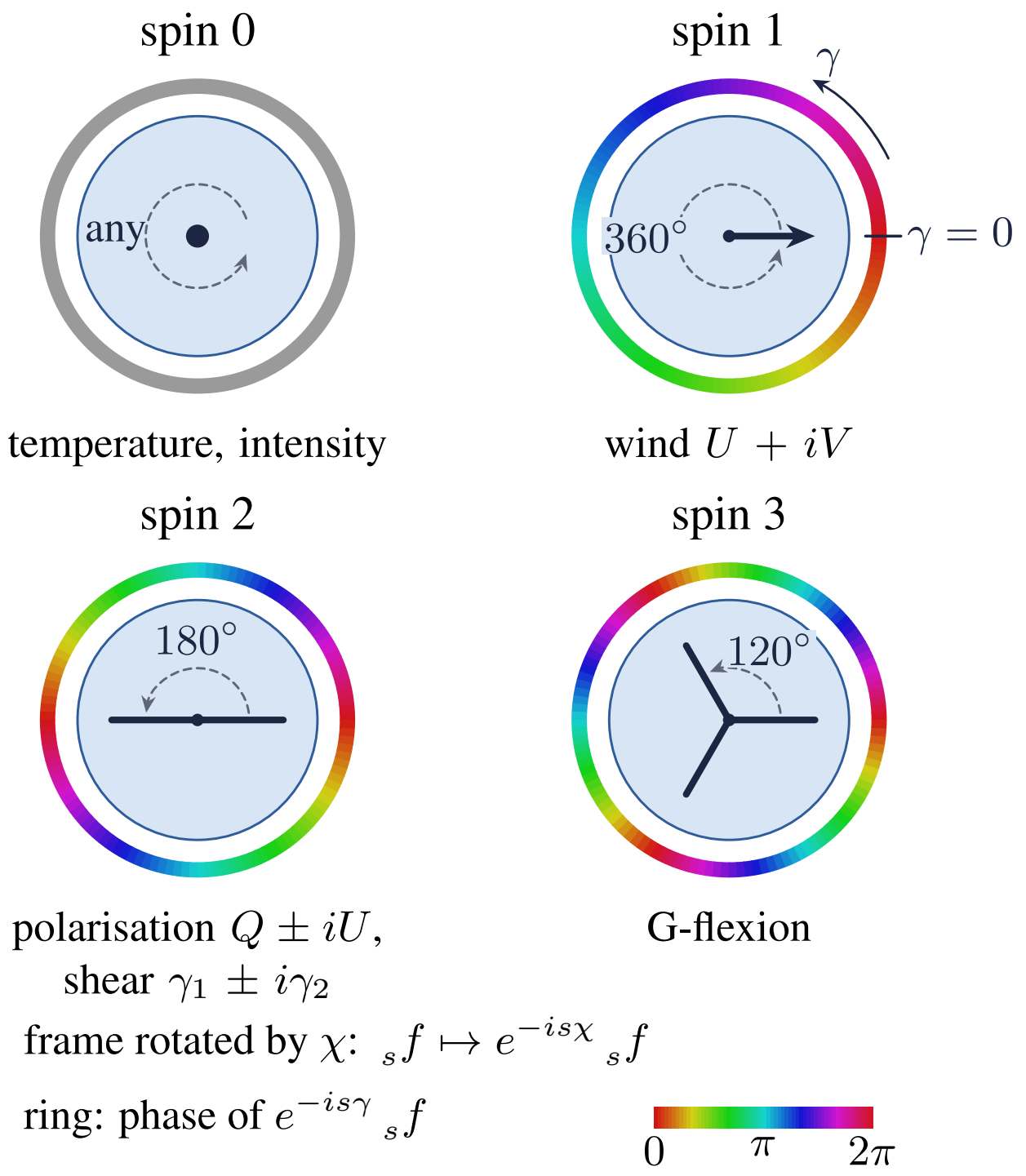}
  \caption{What spin means at a point. Each panel is the tangent plane at $x$, with $e_\theta$ to the right. A spin-$s$ value is drawn as $s$ equal spokes, and rotating the frame by $\chi$ multiplies it by $e^{-\ii s\chi}$, as in \eqref{eq:spin-phase}. The ring, the fibre of $\SOthree$ over $x$ (Sec.~\ref{sec:lifting}), is coloured by the phase of the lift of the field, which winds $s$ times around it.}
  \label{fig:spin-glyphs}
\end{figure}

\begin{figure}[htbp]
  \centering
  \includegraphics[width=\textwidth]{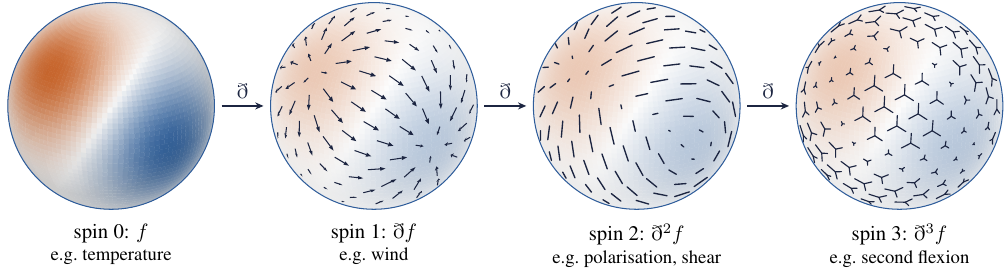}
  \caption{Spin fields on the sphere: a scalar field $f$ and the fields $\eth f$, $\eth^2 f$ and $\eth^3 f$ obtained from it by the spin raising operator $\eth$ of \eqref{eq:eth}, drawn with glyphs whose symmetry is their spin (Fig.~\ref{fig:spin-glyphs}). Wind has spin $1$, polarisation and shear spin $2$, and the second flexion of weak lensing spin $3$ \citep{bacon2006weak}.}
  \label{fig:spin-fields}
\end{figure}

The Wigner $D$-functions, the matrix elements of the irreducible unitary representations of $\SOthree$ \citep{varshalovich1988quantum}, may be written
\begin{equation}
  D^{\ell}_{mn}(\alpha,\beta,\gamma) = e^{-\ii m\alpha}\, d^{\ell}_{mn}(\beta)\, e^{-\ii n\gamma},
  \label{eq:wignerD}
\end{equation}
where the Wigner $d$-functions $d^{\ell}_{mn}$ are real. They satisfy $D^{\ell}(\rho\rho')=D^{\ell}(\rho)D^{\ell}(\rho')$ and $D^{\ell}(\rho^{-1})=D^{\ell}(\rho)^{\dagger}$ and the orthogonality relation $\langle D^{\ell}_{mn},D^{\ell'}_{m'n'}\rangle=\frac{8\pi^2}{2\ell+1}\delta_{\ell\ell'}\delta_{mm'}\delta_{nn'}$, and by the Peter--Weyl theorem every $F\in\Lrot$ admits the Wigner expansion
\begin{equation}
  F(\rho) = \sum_{\ell=0}^{\infty}\frac{2\ell+1}{8\pi^2}\sum_{m,n=-\ell}^{\ell} F^{\ell}_{mn}\,D^{\ell *}_{mn}(\rho),
  \qquad
  F^{\ell}_{mn} = \int_{\SOthree} F(\rho)\, D^{\ell}_{mn}(\rho)\dmu(\rho) .
  \label{eq:wigner-transform}
\end{equation}
The identity tying the sphere to the rotation group is
\begin{equation}
  e^{-\ii s\gamma}\, \sw{s}{Y}_{\ell m}(\beta,\alpha) = (-1)^s \sqrt{\frac{2\ell+1}{4\pi}}\; D^{\ell *}_{m,-s}(\alpha,\beta,\gamma) .
  \label{eq:D-sY}
\end{equation}
Since the lift of a spin-$s$ function thus occupies the Wigner column $n=-s$ and intertwines the spin rotation with left translation (Lemma~\ref{lem:lift}(c) and (e)), and left translation acts on the first Wigner index alone, the harmonic coefficients of a function of any spin rotate by the same matrices under the spin rotation operator $\Rot_\rho$ of \citet{mcewen2015directional},
\begin{equation}
  (\Rot_\rho\, \sw{s}{f})_{\ell m} = \sum\nolimits_{n} D^{\ell}_{mn}(\rho)\, \sw{s}{f}_{\ell n} .
  \label{eq:rotation-coeffs}
\end{equation}
Figs.~\ref{fig:spin-glyphs} and~\ref{fig:spin-fields} show spin at a point and over the sphere.

\subsection{Spin fields, Mackey functions and the lift}
\label{app:lift}

\begin{figure}[t]
  \centering
  \includegraphics[width=0.4\textwidth,page=2]{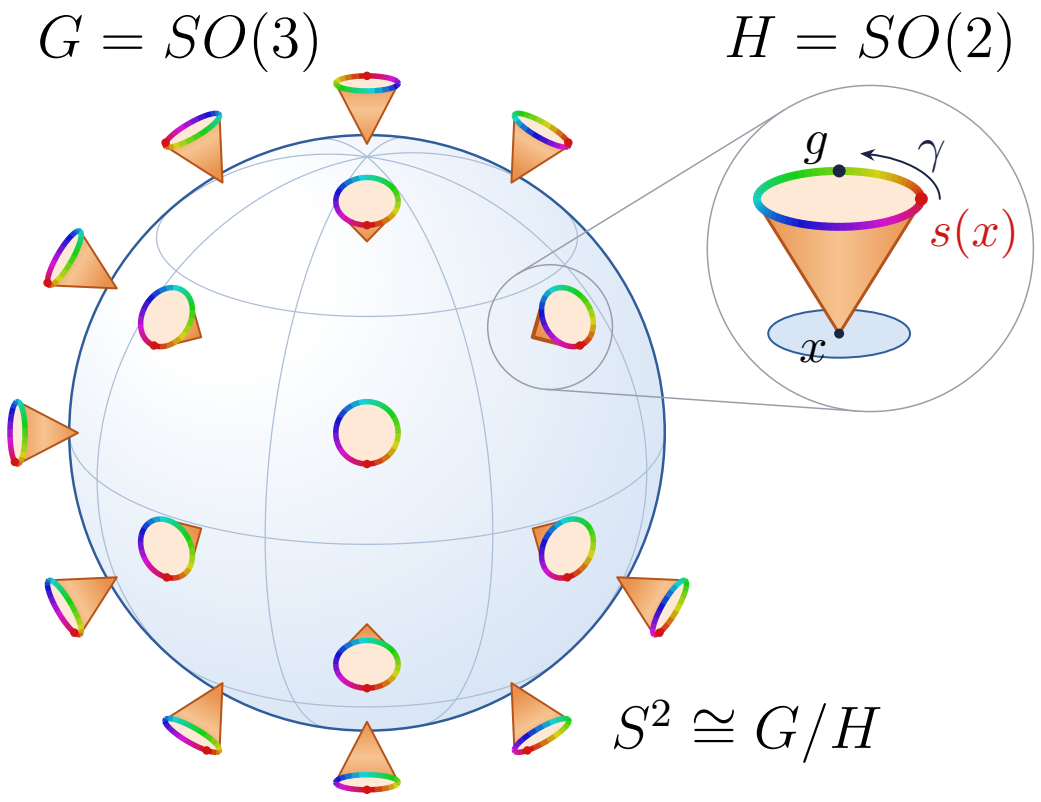}
  \caption{The rotation group as a principal $\SOtwo$-bundle over the sphere. The fibre over $x$, the rotations $R(\alpha,\beta,\gamma)$ with $R(\alpha,\beta,\gamma)\eta = x$, is drawn as the rim of a cone at $x$, coloured by the fibre coordinate $\gamma$, with the section $s(x) = R(\alpha,\beta,0)$ as a red dot and a general element $g = s(x)R_z(\gamma)$ in the inset.}
  \label{fig:bundle}
\end{figure}

The rotation group is a principal $\SOtwo$-bundle over the sphere with projection $p$, whose fibre over a point is the circle of unit tangent vectors there (Fig.~\ref{fig:bundle}). We adopt the section
\begin{equation}
  s(\theta,\varphi) = R(\varphi,\theta,0), \qquad\text{\ie}\qquad (\beta,\alpha) \mapsto (\alpha,\beta,0)
  \label{eq:section}
\end{equation}
for $0<\theta<\pi$, with $s(\eta)=e$ and $s(-\eta)=R_y(\pi)$ at the poles. It selects at each point the frame aligned with the meridian direction $e_\theta$ and is measurable, which suffices since all constructions are in $L^2$. The irreducible representations of $\SOtwo$ are the characters $\rho^{(s)}(R_z(\gamma))=e^{\ii s\gamma}$, $s\in\mathbb{Z}$, and the Mackey functions of type $s$ \citep[Definition~IV.4]{mcewen2026unifying} are the elements of
\begin{equation}
  \Mackey{s} = \{ F \in \Lrot : F(gR_z(\gamma)) = e^{-\ii s\gamma}F(g) \} .
  \label{eq:mackey}
\end{equation}
Changing the section to $s(x)R_z(\chi(x))$ multiplies the coordinate expression $\sqrt{2\pi}\,F(s(x))$ of a Mackey function by $e^{-\ii s\chi(x)}$, which is the frame law \eqref{eq:spin-phase} of a spin-$s$ field \citep[eq.~36]{mcewen2026unifying}.

\begin{definition}[{Lift and projection; \citealp[Definition~IV.6]{mcewen2026unifying}}]
\label{def:lift}
For $s \in \mathbb{Z}$ define the lift $\lift{s} : \Lsph{s} \to \Lrot$ and its adjoint, the projection $\proj{s} : \Lrot \to \Lsph{s}$, by
\begin{align}
  (\lift{s} f)(\alpha,\beta,\gamma) &= \frac{1}{\sqrt{2\pi}}\, e^{-\ii s\gamma}\, f(\beta,\alpha), \label{eq:lift}\\
  (\proj{s} F)(\beta,\alpha) &= \frac{1}{\sqrt{2\pi}} \int_0^{2\pi} e^{\ii s\gamma}\, F(\alpha,\beta,\gamma) \dd\gamma .
  \label{eq:proj}
\end{align}
\end{definition}

\begin{lemma}[{Structure of the lift; \citealp[Lemma~IV.7]{mcewen2026unifying}}]
\label{lem:lift}
Let $s \in \mathbb{Z}$.
\begin{enumerate}\itemsep0pt
  \item[(a)] $\lift{s}$ is an isometry of $\Lsph{s}$ onto $\Mackey{s}$; $\proj{s}\lift{s} = \Id$ and $\lift{s}\proj{s}$ is the orthogonal projector onto $\Mackey{s}$.  The subspaces $\Mackey{s}$ are mutually orthogonal and $\Lrot = \widehat{\bigoplus}_{s \in \mathbb{Z}} \Mackey{s}$.
  \item[(b)] On $\Mackey{s}$ the projection is evaluation on the section: $\proj{s} F = \sqrt{2\pi}\, F \circ s$.
  \item[(c)] In harmonic space,
  \begin{align}
    (\lift{s} f)^{\ell}_{mn} &= (-1)^s \sqrt{\frac{8\pi^2}{2\ell+1}}\; \sw{s}{f}_{\ell m}\, \delta_{n,-s}, \label{eq:lift-harmonic}\\
    (\proj{s} F)_{\ell m} &= (-1)^s \sqrt{\frac{2\ell+1}{8\pi^2}}\; F^{\ell}_{m,-s} ,
    \label{eq:proj-harmonic}
  \end{align}
  where the coefficients on the sphere are taken with respect to the spin-$s$ harmonics.
  \item[(d)] $\lift{s}\, \sw{s}{Y}_{\ell m} = (-1)^s \sqrt{\frac{2\ell+1}{8\pi^2}}\; D^{\ell *}_{m,-s}$: lifted spin harmonics are unit-norm Wigner functions.
  \item[(e)] $\piG(g)\lift{s} = \lift{s}\pis{s}(g)$ and $\proj{s}\piG(g) = \pis{s}(g)\proj{s}$ for all $g \in \SOthree$, where $\piG$ is the left regular representation of $\SOthree$ on $\Lrot$, $(\piG(g)F)(g') = F(g^{-1}g')$, and $\pis{s}$ is the representation of $\SOthree$ on spin-$s$ fields induced from $\rho^{(s)}$; both are developed in \citet[Sec.~IV.D]{mcewen2026unifying}, where (e) is proved as Proposition~IV.10.
\end{enumerate}
\end{lemma}

By \citet[Proposition~IV.10]{mcewen2026unifying}, the induced representation $\pis{s}(g)=\proj{s}\piG(g)\lift{s}$ is the spin rotation operator $\Rot_\rho$ of \citet{mcewen2015directional}.

\begin{corollary}[{Functions on $\SOthree$ as tuples of spin fields; \citealp[Corollary~IV.12]{mcewen2026unifying}}]
\label{cor:spin-tuple}
Every $F \in \Lrot$ decomposes uniquely as $F = \sum_{s \in \mathbb{Z}} \lift{s}\proj{s}F$, its Fourier series in $\gamma$ converging in $L^2$, with $(\lift{s}\proj{s}F)^{\ell}_{mn} = F^{\ell}_{mn}\,\delta_{n,-s}$; the component $\proj{s}F$ is a spin-$s$ field whose harmonic coefficients are the Wigner column $F^{\ell}_{m,-s}$ up to the constant in \eqref{eq:lift-harmonic}.  In particular $F$ is azimuthally band-limited at $N$, $F^{\ell}_{mn} = 0$ for $\lvert n \rvert \ge N$, if and only if $F = \sum_{\lvert s \rvert < N} \lift{s} f_s$ for spin fields $f_s = \proj{s}F$.
\end{corollary}

\subsection{Lifted convolutions}
\label{app:convolutions}

The statement summarised in Sec.~\ref{sec:convolutions}, for the convolutions \eqref{eq:group-conv} and \eqref{eq:spin-conv-def} \citep[Definitions~III.4 and V.1]{mcewen2026unifying}, is the following.

\begin{theorem}[{Harmonic form; \citealp[Theorem~V.2]{mcewen2026unifying}}]
\label{thm:harmonic}
Let $\Psi \in \Lrot$, $s, s' \in \mathbb{Z}$, and let $\widetilde{\Psi}(g) = \Psi^{*}(g^{-1})$.  For $\lvert m\rvert, \lvert n\rvert \le \ell$, with $\ell \ge \lvert s\rvert$ whenever the input is a spin-$s$ field and $\ell \ge \lvert s'\rvert$ whenever the output is a spin-$s'$ field,
\begin{gather}
  \big(\Cop{s}{G}[\Psi]\, \sw{s}{f}\big)^{\ell}_{mn} = (-1)^s \sqrt{\tfrac{8\pi^2}{2\ell+1}}\; \sw{s}{f}_{\ell m}\; \Psi^{\ell *}_{n,-s} ,
  \label{eq:harmonic-sG}\\
  \big(\Cop{s}{s'}[\Psi]\, \sw{s}{f}\big)_{\ell m} = (-1)^{s+s'}\; \sw{s}{f}_{\ell m}\; \Psi^{\ell *}_{-s',-s} ,
  \label{eq:harmonic-ss}\\
  \big(\Cop{G}{s'}[\Psi]\, F\big)_{\ell m} = (-1)^{s'} \sqrt{\tfrac{2\ell+1}{8\pi^2}}\; \sum_{m'=-\ell}^{\ell} F^{\ell}_{mm'}\; \Psi^{\ell *}_{-s',m'} ,
  \label{eq:harmonic-Gs}\\
  \big(\Cop{G}{G}[\Psi]\, F\big)^{\ell}_{mn} = \sum_{m'=-\ell}^{\ell} F^{\ell}_{mm'}\; \Psi^{\ell *}_{nm'} ,
  \label{eq:harmonic-GG}
\end{gather}
and the output of $\Cop{s}{s'}[\Psi]$ has no component of degree $\ell < \max(\lvert s\rvert,\lvert s'\rvert)$.  All four operators are bounded and $\SOthree$-equivariant for every $\Psi$: $\Cop{s}{s'}[\Psi]\,\pis{s}(g) = \pis{s'}(g)\,\Cop{s}{s'}[\Psi]$ and correspondingly with $\piG$ in place of $\pis{s}$ or $\pis{s'}$.  Their adjoints are $\big(\Cop{s}{G}[\Psi]\big)^{\dagger} = \Cop{G}{s}[\widetilde\Psi]$, $\big(\Cop{s}{s'}[\Psi]\big)^{\dagger} = \Cop{s'}{s}[\widetilde\Psi]$ and $\big(\Cop{G}{G}[\Psi]\big)^{\dagger} = \Cop{G}{G}[\widetilde\Psi]$.
\end{theorem}

In outline, the lift of $\sw{s}{f}$ occupies the Wigner column $n=-s$ (Lemma~\ref{lem:lift}(c)), and the convolution \eqref{eq:group-conv-harmonic} multiplies each block of degree $\ell$ on the right by $(\Psi^{\ell})^{\dagger}$, so that only the kernel column $m'=-s$ is read, which gives \eqref{eq:harmonic-sG}. The projection \eqref{eq:proj-harmonic} reads the column $n=-s'$, so that only $\Psi^{\ell}_{-s',-s}$ reaches the output and the $\ell$-dependent constants cancel, which gives \eqref{eq:harmonic-ss}. Equivariance follows from Lemma~\ref{lem:lift}(e) and the fact that the group convolution commutes with left translation, and the full proof is in \citet[Theorem~V.2]{mcewen2026unifying}.

By \eqref{eq:rotation-coeffs} the spaces $\Lsph{s}$ and $\Lsph{s'}$ carry one copy of the irreducible representation $D^{\ell}$ for each $\ell\ge\max(\lvert s\rvert,\lvert s'\rvert)$, so that completeness follows from Schur's lemma.

\begin{theorem}[{Completeness; \citealp[Theorem~V.5]{mcewen2026unifying}}]
\label{thm:completeness}
Let $T : \Lsph{s} \to \Lsph{s'}$ be a bounded complex-linear map with $T\pis{s}(g) = \pis{s'}(g)T$ for all $g \in \SOthree$. Then there is a unique bounded sequence $(c_\ell)_{\ell \ge \max(\lvert s\rvert,\lvert s'\rvert)}$ such that
\begin{equation}
  (T\,\sw{s}{f})_{\ell m} = c_\ell\;\sw{s}{f}_{\ell m} ,
  \label{eq:classification}
\end{equation}
with input degrees absent from the target annihilated. Conversely every bounded sequence defines such a $T$, with $\lVert T\rVert = \sup_\ell\lvert c_\ell\rvert$.
\end{theorem}

\begin{proposition}[{Azimuthally band-limited and steerable kernels; \citealp[Proposition~VI.6]{mcewen2026unifying}}]
\label{prop:directional-tuple}
If the column $m' = -s$ of $\Psi$ is supported on $\lvert n \rvert < N$ then
\begin{equation}
  \Cop{s}{G}[\Psi] = \sum_{\lvert s'\rvert < N} \lift{s'}\, \Cop{s}{s'}[\Psi] ,
  \label{eq:directional-tuple}
\end{equation}
\ie\ the output is represented by at most $2N-1$ spin-changing components $\Cop{s}{s'}[\Psi]$, $\lvert s'\rvert<N$, with $2\min(\ell,N-1)+1$ admissible output slots at degree $\ell$.  Conversely, any finite tuple of bounded equivariant spin-changing maps from the same input spin assembles into a bounded map $\Cop{s}{G}$ with a generally distributional kernel.  A square integrable representative exists exactly when the sum of their Hilbert--Schmidt norms squared is finite.  At finite band-limit the assembled kernel is always square integrable.
\end{proposition}

\begin{corollary}[{Incompatibility of directional kernels, spherical output and equivariance; \citealp[Corollary~V.7]{mcewen2026unifying}}]
\label{cor:nogo}
For a bounded complex-linear equivariant map $T : \Lsph{s} \to \Lsph{s'}$ between one fixed input spin and one fixed output spin, the two-point kernel is
\begin{equation}
  K(x,y)=(-1)^{s+s'}\sum_\ell c_\ell\frac{2\ell+1}{4\pi}
  D^\ell_{-s',-s}\big(s(x)^{-1}s(y)\big),
  \label{eq:nogo-kernel}
\end{equation}
in the distributional sense in general. It is radial apart from the frame phases prescribed by the two spins. Directional kernels, output on the sphere and $\SOthree$-equivariance therefore cannot be achieved simultaneously beyond those prescribed phases for this input/output pair.
\end{corollary}

\section{Spherical layers as neural operators}
\label{app:lineage}

Table~\ref{tab:lineage} summarises the spherical layers of Sec.~\ref{sec:background:operators} as neural operators, with the year of the first public version of each work (the FNO and the generalised spherical CNN appeared in the same month), although resolution-independent parameters imply neither exact sampled non-linear equivariance nor learned weights at unseen degrees.

\begin{table}[htbp]
  \caption{Spherical convolutional layers as neural operators. A layer is a neural operator in the sense of \citet{kovachki2023neural} when its parameters do not depend on the sampling of the sphere. Representative costs at bandlimit $L$ (azimuthal bandlimit $N\le L$, $n$ planar grid points) assume fixed channel and retained spin counts and, for DISCO, bounded local neighbourhood size.}
  \label{tab:lineage}
  \centering\small
  \begin{tabular}{lllll}
    \toprule
    Method & Year & Signals & Linear layer & Cost \\
    \midrule
    Spherical CNN \citep{esteves2018learning} & 2018 & $\sphere$, scalar & harmonic zonal filter, one weight per $\ell$ & $L^3$ \\
    Spherical CNN \citep{cohen2018spherical} & 2018 & $\SOthree$, scalar & grid filter via generalised FFT & $L^4$ \\
    Clebsch--Gordan net \citep{kondor2018clebsch} & 2018 & $\SOthree$, harmonic & per-$\ell$ maps, CG products & $L^5$ \\
    Spin-weighted CNN \citep{esteves2020spin} & 2020 & $\sphere$, spins & harmonic filters interpolated over $\ell$ & $L^3$ \\
    Generalised spherical CNN \citep{cobb2021efficient} & 2020 & $\sphere$, $\SOthree$ & per-$\ell$ maps, azimuthal bandlimit $N$ & $NL^3$ \\
    FNO \citep{li2021fourier} & 2020 & plane & one multiplier per mode & $n\log n$ \\
    DISCO \citep{ocampo2023scalable} & 2022 & $\sphere$, scalar & continuous filter, quadrature & $L^2$ \\
    SFNO \citep{bonev2023spherical} & 2023 & $\sphere$, scalar & one weight per $\ell$ & $L^3$ \\
    \SSFNO{} (this work) & 2026 & $\sphere$, $\SOthree$, spins & real profile in $\ell$ per spin pair (10 knots) & $L^3$ \\
    \bottomrule
  \end{tabular}
\end{table}

\section{Equivariant non-linearities}
\label{app:nonlinearity}

We consider the two settings of Sec.~\ref{sec:nonlinearity} in turn, the sphere and the rotation group.

\subsection{Non-linearities on the sphere}
\label{app:nonlinearity:sphere}

A rotation moves a spin-$s$ field and multiplies its value at each point by a phase $e^{-\ii s\chi}$ of the form \eqref{eq:spin-phase}, where $\chi$ is the angle by which the local frame turns there. A pointwise map $\zeta:\mathbb{C}\to\mathbb{C}$ commutes with the move and is therefore equivariant if and only if it commutes with every such phase, which for $s\neq0$ reads $\zeta(e^{\ii\phi}z)=e^{\ii\phi}\zeta(z)$ for all $\phi$ and $z$, since the rotations about each point turn its frame by every angle. The modulus non-linearity \eqref{eq:modrelu} multiplies $z$ by a real function of $\lvert z\rvert$, so that
\begin{equation}
  \zeta(e^{\ii\phi}z) = \mathrm{ReLU}\big(a\lvert z\rvert+b\big)\,\frac{e^{\ii\phi}z}{\lvert z\rvert+\epsilon} = e^{\ii\phi}\,\zeta(z),
  \label{eq:modrelu-phase}
\end{equation}
and it is therefore equivariant for every spin. For scalar fields ($s=0$) every pointwise map is equivariant and standard activation functions apply.

\subsection{Non-linearities on the rotation group}
\label{app:nonlinearity:group}

On $\SOthree$ a rotation only translates the argument of a function, so that every pointwise map commutes with it. The lift-and-sum $\liftsum$ places a tuple of spin fields, one for each distinct spin within a channel, in one function on $\SOthree$, each in its own Wigner column by Lemma~\ref{lem:lift}(c), a pointwise map $\zeta$ acts on that function, and a return $\mathcal{P}$ of projecting convolutions maps the result back to a tuple of spin fields. Definition~\ref{def:nonlinearity} states this construction and Theorem~\ref{thm:nonlinearity} its equivariance.

\begin{definition}[Lifted non-linearity]
\label{def:nonlinearity}
For finite input and output spin sets, indexed by $(s_i)$ and $(s'_j)$, kernels $\Psi^j\in\Lrot$ and a measurable pointwise map $\zeta:\mathbb{C}\to\mathbb{C}$ satisfying $\lvert\zeta(z)\rvert\le A+B\lvert z\rvert$ for finite nonnegative constants $A,B$, let
\begin{equation}
  \sigma = \mathcal{P} \circ \zeta \circ \liftsum, \qquad \mathcal{P} = \bigoplus_j \Cop{G}{s'_j}[\Psi^j], \qquad \liftsum\Big(\bigoplus_i \sw{s_i}{f}\Big) = \sum_i \lift{s_i}\,\sw{s_i}{f} .
  \label{eq:nonlinearity}
\end{equation}
\end{definition}

\begin{theorem}[Equivariance of the lifted non-linearity; \citealp{xu2022unified}]
\label{thm:nonlinearity}
Under the hypotheses of Definition~\ref{def:nonlinearity}, $\sigma$ is a well-defined map between the corresponding finite sums of spin $L^2$ spaces and
\begin{equation}
  \sigma \circ \bigoplus_i \pis{s_i}(g) = \bigoplus_j \pis{s'_j}(g) \circ \sigma \qquad \text{for all } g \in \SOthree .
  \label{eq:nonlinearity-equivariance}
\end{equation}
\end{theorem}

Theorem~\ref{thm:nonlinearity} is the result of \citet[Sec.~3.4]{xu2022unified}, who show that their lift-and-sum, Eq.~(3), followed by an elementwise non-linearity and a projection satisfies their condition, Eq.~(2), proved in the same way. The growth condition is ours.

\begin{proof}[Proof of Theorem~\ref{thm:nonlinearity}]
The finite sum $\liftsum$ is bounded into $L^2(\SOthree)$. The growth hypothesis gives $\lVert\zeta(F)\rVert_2\le A\sqrt{8\pi^2}+B\lVert F\rVert_2$, and each component of $\mathcal{P}$ is bounded, so the composition is well defined. With the notation of Definition~\ref{def:nonlinearity},
\begin{align}
  \sigma\circ\bigoplus_i\pis{s_i}(g)
  &= \mathcal{P}\,\zeta\,\liftsum\bigoplus_i\pis{s_i}(g) \notag\\
  &= \mathcal{P}\,\zeta\,\piG(g)\,\liftsum && \text{since } \textstyle\liftsum\bigoplus_i\pis{s_i}(g) = \sum_i\lift{s_i}\pis{s_i}(g) = \sum_i\piG(g)\lift{s_i} = \piG(g)\liftsum \text{ by Lemma~\ref{lem:lift}(e)} \notag\\
  &= \mathcal{P}\,\piG(g)\,\zeta\,\liftsum && \text{since } \zeta \text{ acts pointwise and } \piG(g) \text{ only translates the argument} \notag\\
  &= \bigoplus_j\pis{s'_j}(g)\,\mathcal{P}\,\zeta\,\liftsum && \text{by Theorem~\ref{thm:harmonic} applied to each } \Cop{G}{s'_j}[\Psi^j] \notag\\
  &= \bigoplus_j\pis{s'_j}(g)\circ\sigma .
\end{align}
\end{proof}

In our networks each component of the return is the fixed projection $\proj{s'}$, read as the Fourier column $n=-s'$ of the sampled group signal, followed by a learned filter for each channel and spin (Appendix~\ref{app:implementation}). By \eqref{eq:proj-harmonic} this projection is the projecting convolution \eqref{eq:harmonic-Gs} with the kernel row $\Psi^{\ell *}_{-s',m'}=\delta_{m',-s'}$, a distributional kernel that lies in $\Lrot$ once truncated at the bandlimit. The filter multiplies this row by one value for each degree and channel, and the constants of the implemented lift-and-sum and Fourier read can be absorbed into $\zeta$ and the filter, so that Theorem~\ref{thm:nonlinearity} covers the non-linearity of our networks in the continuum, and the group normalisation between $\zeta$ and the return uses invariant statistics and so preserves equivariance (Sec.~\ref{sec:error-sources}). Xu's fixed fibre integral is $\sqrt{2\pi}\proj{s'}$ with their unnormalised stabiliser measure \citep{xu2022unified}, and the orthogonal group projection of \citet{ballerin2025so3} is $\lift{s'}\proj{s'}$.

\subsection{Discretisation}
\label{app:nonlinearity:discretisation}

Theorem~\ref{thm:nonlinearity} concerns continuum functions, and $\zeta(F)$ is generally not bandlimited even when $F$ is. The exact truncation $Q_{L,N}$ to the Wigner coefficients with $\ell<L$ and $\lvert n\rvert<N$ commutes with rotations, which act only on $m$, whereas computing these coefficients from samples aliases the content of $\zeta(F)$ beyond the bandlimits into them, which oversampling reduces \citep{cobb2021efficient}. Apart from the Conv classifiers, which oversample by a factor of two (Appendix~\ref{app:implementation}), our networks evaluate $\zeta$ without oversampling.

\subsection{Parity}
\label{app:parity}

The antipodal map $P:x\mapsto-x$ commutes with rotations and multiplies scalar harmonic coefficients by $(-1)^\ell$. Networks on scalar fields built from isotropic multipliers and pointwise maps are therefore $\mathrm{O}(3)$-equivariant \citep{weiler2021coordinate} and cannot represent a map that anticommutes with $P$, $T(f\circ P)=-T(f)\circ P$, such as the advection $J(\psi,\zeta)=\hat r\cdot(\nabla\psi\times\nabla\zeta)$ of a vorticity $\zeta=\nabla^2\psi$ treated as a scalar. This is the counterpart, for such networks, of the selection rule of Gaunt tensor products \citep{xie2025price}. Spin-weighted and vector spherical signals restore the couplings of odd total degree \citep{liang2026spin,xie2026asymptotically}, and spin fields escape the restriction, since $P$ acts on a spin-$s$ field by $f\mapsto(-1)^sf^{*}\circ P$. The vorticity is carried by the imaginary part of $\bar\eth(v_\theta+\ii v_\varphi)$, and one spin-$1$ stage represents its advection through $(\eth\psi)^{*}\eth\zeta=\nabla\psi\cdot\nabla\zeta+\ii J(\psi,\zeta)$ with an appropriate pointwise map on $\SOthree$ that is not odd, unlike the modulus non-linearity. A multiplier between spins commutes with $P$, equivalently with the mirror $f\mapsto f^{*}(\theta,-\varphi)$, if and only if its coefficients are real, the spin counterpart of the fixed phase of reflection-equivariant planar kernels \citep{weiler2019general}. The opposite parities of the $E$ and $B$ modes \citep{zaldarriaga1997all} are an instance of this criterion, which a mirror test of our classifiers with real profiles confirms. Completeness among real-linear maps requires the conjugate field $f^{*}$, as the conjugate frequencies do for planar harmonic networks \citep{weiler2019general}.

\section{Network architectures}
\label{app:implementation}

We describe the layers and networks of \texttt{s2ai} used for the reported results, with the notation and names of Sec.~\ref{sec:architecture}.

\subsection{Layers}

\begin{figure}[htbp]
  \centering
  \includegraphics[width=\textwidth]{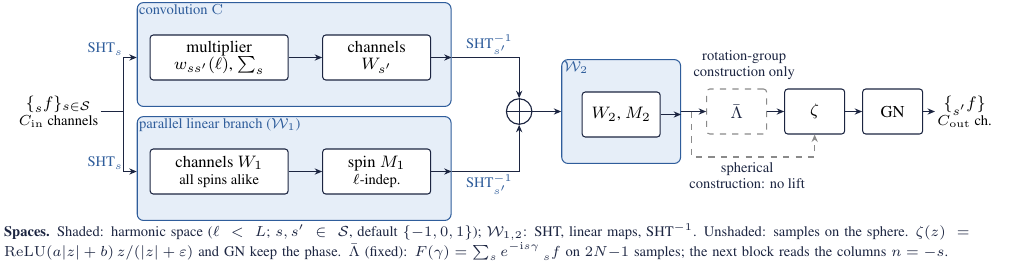}
  \caption{An \SSFNO{} block as implemented. The convolution $\mathrm{C}$, here the spin spherical convolution \eqref{eq:implemented-conv}, and a harmonic $1\times1$ convolution ($\mathcal{W}_1$) with one channel matrix and one spin matrix shared across degrees act in parallel, and their sum passes through a second such convolution ($\mathcal{W}_2$), the modulus non-linearity $\zeta$ and group normalisation (GN). Only the rotation-group construction forms the lift-and-sum $\liftsum$.}
  \label{fig:architecture}
\end{figure}

\paragraph{Spin spherical convolution.} The spherical layer computes the forward spin-$s$ transform of every input channel and spin, truncates or zero-pads the coefficients to the output bandlimit $L_{\rm out}$, and forms for each input channel $c$ and output spin $s'$
\begin{equation}
  \sw{s'}{g}^{\,c}_{\ell m} = \sqrt{\tfrac{4\pi}{2\ell+1}}\,\sum_{s\in\spinset_{\rm in}} w^{c}_{ss'}(\ell)\;\sw{s}{f}^{\,c}_{\ell m} ,
  \label{eq:implemented-conv}
\end{equation}
followed by a real channel-mixing matrix $W_{s'}$ for each output spin and inverse spin-$s'$ transforms, so that the multipliers of the layer form a real and parameter-efficient subfamily of those of Theorem~\ref{thm:completeness}, interpolated in $\ell$. Each real profile $w^{c}_{ss'}(\ell)$, independent of the output channel, is the piecewise-linear interpolation of $10$ learned values at knots evenly spaced on $[0,L_{\rm out}]$. The profile values are initialised Glorot-uniformly with fans $C_{\rm in}\lvert\spinset_{\rm in}\rvert\lvert\spinset_{\rm out}\rvert$ and $10$, and the channel matrices $W_{s'}$, of shape $[C_{\rm in},C_{\rm out},\lvert\spinset_{\rm out}\rvert]$, Glorot-uniformly with the default fan axes of JAX. Neither has a bias. The factor $\sqrt{4\pi/(2\ell+1)}$ is that of evaluation on the section, and the factor $(-1)^{s+s'}$ and the conjugation of \eqref{eq:spin-multiplier} are absorbed into the real weights. Fig.~\ref{fig:architecture} shows an \SSFNO{} block as implemented.

\paragraph{Lift-and-sum and projecting convolution.} The lift-and-sum, which has no parameters, maps a tuple $(\sw{s}{f})_{s\in\spinset}$ to the samples $F(\gamma_k,x)=\sum_{s\in\spinset}e^{-\ii s\gamma_k}\,\sw{s}{f}(x)$ at $\gamma_k=2\pi k/(2N-1)$, $k=0,\dots,2N-2$, without the factor $(2\pi)^{-1/2}$ of Definition~\ref{def:lift}. The projecting layer reads the columns $n=-s$, $s\in\spinset$, of a group signal by the discrete Fourier sum $(2N-1)^{-1}\sum_kF(\gamma_k,x)\,e^{\ii s\gamma_k}$, takes the forward spin-$s$ transform of each, truncates or pads, multiplies by $\sqrt{4\pi/(2\ell+1)}\,w^{c}_{s}(\ell)$ with one real profile per channel and spin, mixes channels within each spin and applies the inverse spin-$s$ transform. Spin $s$ is thus mapped to spin $s$ only, by the fixed adjoint up to a constant followed by a learned spin-preserving filter.

\begin{figure*}[t]
  \centering
  \includegraphics{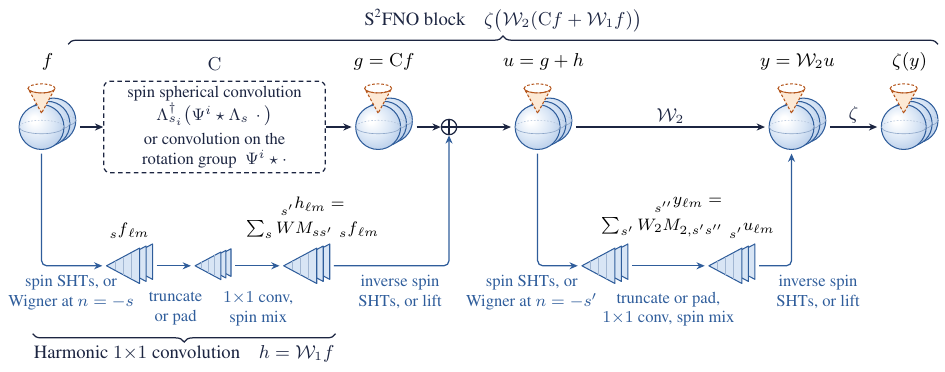}
  \caption{The generic \SSFNO{} block \eqref{eq:spinsfno-block} in detail. Stacked spheres with a dashed cone denote a tuple of spin fields or its lift to $\SOthree$, and the normalisation is not shown.}
  \label{fig:spinsfno-block-detail}
\end{figure*}

\paragraph{Harmonic $1\times1$ convolutions.} The harmonic $1\times1$ convolutions $\mathcal{W}_1$ and $\mathcal{W}_2$ of the \SSFNO{} blocks (Fig.~\ref{fig:spinsfno-block-detail}) mix channels by one real matrix of shape $[C_{\rm in},C_{\rm out}]$ shared by all spins, and spins by one real matrix of shape $[\lvert\spinset_{\rm in}\rvert,\lvert\spinset_{\rm out}\rvert]$ shared by all degrees, orders and channels, both initialised Glorot-uniformly and without bias. The spin matrix gives spin-changing multipliers constant in $\ell$, so that the stage is equivariant.

\paragraph{Activation and normalisation.}\label{app:normalisation} The activation \eqref{eq:modrelu} has a real gain and offset for each channel, initialised to $1$ and $0$, so that it is the identity up to $\epsilon=10^{-8}$ at initialisation. Group normalisation splits the channels into groups and, for each sample and group, computes the quadrature-weighted mean $\mu$ of the complex field and the mean squared modulus $\sigma^2$ of $x-\mu$. It returns $\big(a\lvert x-\mu\rvert/\sqrt{\sigma^2+\epsilon}+b\big)\,x/(\lvert x\rvert+\epsilon)$, with real $a$ and $b$ for each group initialised to $1$ and $0$, so that only the modulus is normalised and the phase of $x$ is kept. Its statistics are computed for each spin separately on the sphere and pooled over the $\gamma$ samples on the group. The pooled mean on the group is the $n=0$ column, a spin-$0$ quantity, and hence an invariant constant, whereas the mean on the sphere of a field of non-zero spin is not rotation invariant, so that the mean subtraction on the sphere is not exactly equivariant for $s\neq0$. A normalisation with statistics computed in harmonic space and no mean subtracted for $s\neq0$ would be exactly equivariant (Sec.~\ref{sec:error-sources}), but the group normalisation already gives reasonable, if not exact, rotational equivariance, with errors of order $10^{-2}$ on the sphere and $10^{-3}$ on the rotation group (Table~\ref{tab:equivariance}).

\subsection{Blocks and spin spherical MNIST networks}

\paragraph{Blocks.} Writing $\mathrm{C}$ for the spin spherical convolution, $\mathrm{C}_G$ for the projecting convolution, $\liftsum$ for the lift-and-sum, $\mathcal{W}_1$ and $\mathcal{W}_2$ for the harmonic $1\times1$ convolutions and $\mathrm{N}$ for the normalisation, the blocks of the four families are
\begin{align*}
  \text{Conv-$\sphere$:}\quad & x\mapsto \mathrm{N}\,\zeta\,\mathrm{C}\,x, &
  \text{Conv-$\SOthree$:}\quad & x\mapsto \mathrm{N}\,\zeta\,\liftsum\,\mathrm{C}_G\,x,\\
  \text{\SSFNO{}-$\sphere$:}\quad & x\mapsto \mathrm{N}\,\zeta\,\mathcal{W}_2\,(\mathrm{C}\,x+\mathcal{W}_1x), &
  \text{\SSFNO{}-$\SOthree$:}\quad & x\mapsto \mathrm{N}\,\zeta\,\mathcal{W}_2\,(\liftsum\,\mathrm{C}_G\,x+\mathcal{W}_1x),
\end{align*}
and no block has an identity residual connection. The first layer of a Conv-$\SOthree$ network is a spin spherical convolution followed by $\liftsum$, $\zeta$ and $\mathrm{N}$, and the first block of a \SSFNO{}-$\SOthree$ network is a \SSFNO{}-$\sphere$ block without activation and normalisation, followed by $\liftsum$, $\zeta$ and $\mathrm{N}$. In the UNets the input layer maps the data spin, $0$ for images and $1$ for vector fields, to $\spinset=\{-1,0,1\}$, with $N=8$ in the rotation-group construction.

\paragraph{Classifiers.} The classifiers of Table~\ref{tab:mnist-classification} follow \citet{esteves2020spin} in their normalisation, downsampling and learning-rate schedule. Each stacks six blocks of a convolution, a normalisation and an activation, with $1\to4\to4\to8\to8\to16\to16$ channels for Conv-$\sphere$, $1\to8\to8\to16\to16\to32\to32$ for Conv-$\SOthree$ and \SSFNO{}-$\sphere$ and $1\to4\to8\to16\to16\to32\to32$ for \SSFNO{}-$\SOthree$, internal spins $\{0,1\}$ in the spherical construction and azimuthal bandlimit $N=4$ in the rotation-group construction. The bandlimit falls from $32$ to $16$ and $8$ over successive pairs of blocks, and the Conv classifiers evaluate their activations on grids oversampled by a factor of two against aliasing. In the Conv classifiers each pair of channels has its own complex profile, and in the \SSFNO{} classifiers each input channel has a complex profile for each pair of spins, followed by channel matrices. The batch normalisation takes statistics over the batch and the quadrature on the sphere (and $\gamma$ on the group) and subtracts the mean only from spin-$0$ fields and on the group, so that it is equivariant. The activation is $\mathrm{ReLU}(\mathrm{Re}\,x+b)$, with a bias $b$ for each channel, on spin-$0$ fields and on the samples on the group, and the modulus non-linearity $\mathrm{ReLU}(\lvert x\rvert+b)\,x/\lvert x\rvert$, with $b$ initialised to $-0.5$, on fields of non-zero spin. The pooling returns only rotation-invariant features, the real and imaginary parts of the mean of the spin-$0$ field, or of the function on the group, and the mean modulus of every field. A linear layer to $128$ units, a Swish activation \citep{ramachandran2017searching} and a linear layer then give the $10$ class scores.

\begin{figure}[htbp]
  \centering
  \includegraphics[width=0.5\textwidth]{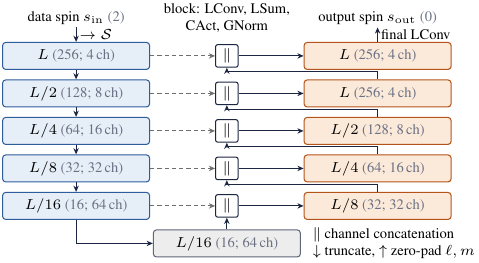}
  \caption{Composition of blocks into the UNets used for dense prediction. Every box is a block of a lifting convolution (LConv), the lift-and-sum (LSum), the activation (CAct) and group normalisation (GNorm). The first block maps the data spin onto the internal spin set $\spinset$ and the last layer returns the output spin. Brackets give the five-level spin UNet of the mass-mapping benchmark (bandlimit; channels), whereas the MNIST UNets are shallower and the \SSFNO{} UNets keep $L=32$ at every level.}
  \label{fig:unet}
\end{figure}

\paragraph{UNets.} The MNIST UNets (Fig.~\ref{fig:unet}) have an encoder of four levels with $8$, $16$, $32$ and $32$ channels and a decoder of two levels with $16$ and $8$ channels, each decoder level concatenating the previous output with the normalised encoder features of equal bandlimit and spin set. In the Conv UNets the bandlimit is $32$ at the first two levels and $16$ at the last two, and is restored to $32$ in the decoder, by truncating and zero-padding the harmonic coefficients inside the convolutions, whereas the \SSFNO{} UNets keep $L=32$ throughout. The final layer, which maps the concatenation of $8+8$ channels to one channel of the target spin, is a spin spherical convolution without activation in Conv-$\sphere$, a projecting convolution in the rotation-group construction and a \SSFNO{}-$\sphere$ block, with its activation and normalisation, in \SSFNO{}-$\sphere$.

\subsection{Networks for wind forecasting and weak lensing}
\label{app:architectures}

For wind forecasting the spin UNet, built from Conv-$\SOthree$ blocks like that of mass mapping, has a lifting input layer that maps the three complex input fields onto the spins $\{-1,0,1\}$, and the scalar UNet takes the real and imaginary parts of each input field as separate channels, six input channels in place of three.

\begin{figure*}[t]
  \centering
  \includegraphics{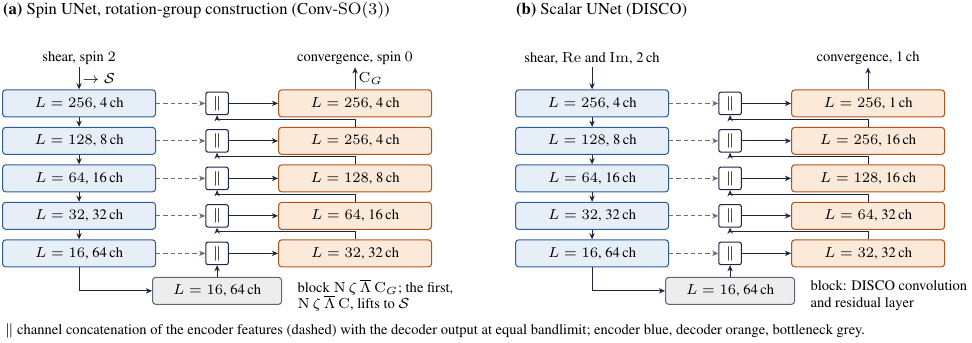}
  \caption{The two mass-mapping UNets, with five levels from $L=256$ to $L=16$. Each box gives the bandlimit and the number of channels of a block, and the dashed lines carry the encoder features, which are concatenated ($\Vert$) with the decoder output at equal bandlimit. (a) The spin UNet of the rotation-group construction, whose first block lifts the spin-$2$ shear to the internal spin set $\spinset$ and whose last layer, a projecting convolution $\mathrm{C}_G$, returns the spin-$0$ convergence. (b) The scalar UNet, of DISCO convolutions and residual layers, which takes the real and imaginary parts of the shear as two channels and returns the convergence as its single channel.}
  \label{fig:lensing-unets}
\end{figure*}

The two mass-mapping UNets (Fig.~\ref{fig:lensing-unets}) have five levels that halve the bandlimit from $L=256$ to $L=16$, with encoders of $4$, $8$, $16$, $32$ and $64$ channels. In the spin UNet the first layer lifts the spin-$2$ shear to the internal spin set, every subsequent block acts on the lifted features, and the final layer returns the spin-$0$ convergence, whereas the scalar UNet combines DISCO convolutions \citep{ocampo2023scalable} with residual layers \citep{he2016deep}, except in its last block, a residual layer alone, and its bottleneck adds a convolution from $128$ to $64$ channels.

\subsection{Parameter counts}

Table~\ref{tab:params} counts trainable parameters only, and its MNIST counts are equal for spin-$0$ and spin-$1$ inputs.

\begin{table}[htbp]
  \caption{Trainable parameter counts of the architectures evaluated in Sec.~\ref{sec:experiments}.}
  \label{tab:params}
  \centering
  \begin{tabular}{llr}
    \toprule
    Task & Architecture & Trainable parameters \\
    \midrule
    Classification & Conv-$\sphere$ & 49{,}674 \\
    Classification & Conv-$\SOthree$ & 133{,}562 \\
    Classification & \SSFNO{}-$\sphere$ & 182{,}016 \\
    Classification & \SSFNO{}-$\SOthree$ & 136{,}088 \\
    \midrule
    Prediction     & Conv-$\sphere$ & 23{,}330 \\
    Prediction     & Conv-$\SOthree$ & 13{,}890 \\
    Prediction     & \SSFNO{}-$\sphere$ & 29{,}097 \\
    Prediction     & \SSFNO{}-$\SOthree$ & 20{,}034 \\
    \midrule
    Wind forecasting & Spin UNet     & 48{,}446 \\
    Wind forecasting & Scalar UNet & 74{,}398 \\
    \midrule
    Mass mapping & Spin UNet     & 80{,}702 \\
    Mass mapping & Scalar UNet & 93{,}198 \\
    \bottomrule
  \end{tabular}
\end{table}

\section{Equivariance tests}
\label{app:equivariance-error}

We measure equivariance by the round-trip test of \citet[Appendix~D]{cobb2021efficient}, also adopted by \citet{ocampo2023scalable}. We consider $N_{\mathrm{sig}}=100$ test signals $\{f_i\}_{i=1}^{N_{\mathrm{sig}}}$ (spin-$s$ signals on the sphere or signals on the rotation group) with complex Gaussian random harmonic coefficients bandlimited at $L$, and $N_{\mathrm{rot}}=100$ rotations $\{\rho_j\}_{j=1}^{N_{\mathrm{rot}}}$ drawn from the Haar distribution on $\SOthree$. For each pair $(f_i,\rho_j)$ we rotate a copy of the coefficients exactly in harmonic space by \eqref{eq:rotation-coeffs}, synthesise both copies on the sampling grid of \citet{mcewen2011novel}, without oversampling, and pass them through the operator under test. We then transform both outputs to harmonic space, rotate the output from the unrotated input by the same rotation, transform both back to real space and compute the mean relative equivariance error
\begin{equation}
  \mraee = \frac{1}{N_{\mathrm{sig}}}\frac{1}{N_{\mathrm{rot}}}\sum_{i=1}^{N_{\mathrm{sig}}}\sum_{j=1}^{N_{\mathrm{rot}}}\frac{\lVert f_{ij} - g_{ij}\rVert_2}{\lVert g_{ij}\rVert_2},
  \label{eq:mraee}
\end{equation}
where $f_{ij}$ denotes the rotated output obtained from the unrotated input, $g_{ij}$ the output obtained from the rotated input, and $\lVert\cdot\rVert_2$ is the $L^2$ norm, as in \citet{cobb2021efficient}, computed over samples, channels and spins or, for group signals, also the $\gamma$ samples, with the colatitude quadrature weights of the sampling theorem. All errors are computed in double precision, at $L=16$ with $N=5$ on the rotation group, except for the linear models of Table~\ref{tab:equiv-linear}, at $L=32$.

The non-linear rows of Table~\ref{tab:equivariance} are tested on spin-$0$ signals on the sphere and on group signals, for which every pointwise map commutes with rotations in the continuum, so that the errors of the activations arise from aliasing. The rows ReLU and Swish (complex data) apply these functions directly to complex values, the ReLU keeping $z$ when $\operatorname{Re}z>0$ and setting it to zero otherwise. The complex activations instead act on the modulus and keep the phase. The complex ReLU is the modulus non-linearity \eqref{eq:modrelu} at its initialisation, $a=1$ and $b=0$, where it is the identity up to $\epsilon=10^{-8}$, so that its error of order $10^{-9}$ is that of the round trip, and it would grow once $a$ and $b$ are learned. Complex Swish is the same non-linearity with the Swish $x/(1+e^{-x})$ \citep{ramachandran2017searching} in place of the ReLU, with the same initial gain and offset. Equivariance of the trained networks, with their learned activations and normalisations, is not measured.

We also test blocks and whole UNets with several spins and channels, with every activation and normalisation removed so that every model is linear (Table~\ref{tab:equiv-linear}). The UNets with $c=8$ and $N=8$ have the architecture of our spin spherical MNIST UNets and are tested in both directions, $0\to1$ and $1\to0$. The identity maps give the round-off floor of the transforms, of order $10^{-14}$ on the sphere and $10^{-13}$ on the rotation group. The positive control, a layer that adds one to the samples in the northern hemisphere, gives an error of order $10^{-1}$, which shows that the test detects broken equivariance. Every linear block and UNet gives errors of order $10^{-14}$, the floor on the sphere, as expected.

\begin{table}[htbp]
\centering
\footnotesize
\setlength{\tabcolsep}{2.5pt}
\begin{tabular}{lcllrc}
\toprule
Model & $N$ & Spins & Channels & Params & Equivariance error ($\mraee$) \\
\midrule
Identity on $\sphere$ & & $1$ & $1$ & $0$ & $2.20\times10^{-14}$ \\
Identity on $\SOthree$ & $8$ & & $1$ & $0$ & $1.99\times10^{-13}$ \\
Positive control & & $1$ & $1$ & $0$ & $8.03\times10^{-2}$ \\
\midrule
\SSFNO{}-$\sphere$ block & & $0{\to}0{\to}0$ & $1,1,1$ & $15$ & $2.27\times10^{-14}$ \\
\SSFNO{}-$\sphere$ block & & $1{\to}1{\to}1$ & $1,1,1$ & $15$ & $2.28\times10^{-14}$ \\
\SSFNO{}-$\sphere$ block$^{\ddagger}$ & & $0{\to}\{-1,0,1\}{\to}\{-1,0,1\}$ & $1,8,8$ & $138$ & $2.00\times10^{-14}$ \\
\SSFNO{}-$\sphere$ block & & $\{-1,0,1\}$ & $8,16,16$ & $1{,}506$ & $2.14\times10^{-14}$ \\
\SSFNO{}-$\sphere$ block & & $\{-2,\dots,2\}$ & $4,4,4$ & $1{,}162$ & $2.07\times10^{-14}$ \\
\SSFNO{}-$\SOthree$ block & $5$ & $\{2\}$ & $1,1,1$ & $15$ & $2.38\times10^{-14}$ \\
\SSFNO{}-$\SOthree$ block & $8$ & $\{-1,0,1\}$ & $8,16,16$ & $1{,}026$ & $2.35\times10^{-14}$ \\
\SSFNO{}-$\SOthree$ block & $8$ & $\{-2,\dots,2\}$ & $4,4,4$ & $362$ & $2.28\times10^{-14}$ \\
\midrule
\SSFNO{}-$\sphere$ UNet & & $0{\to}\{-1,0,1\}{\to}1$ & $c=8$ & $28{,}825$ & $2.02\times10^{-14}$ \\
\SSFNO{}-$\sphere$ UNet & & $1{\to}\{-1,0,1\}{\to}0$ & $c=8$ & $28{,}825$ & $1.98\times10^{-14}$ \\
\SSFNO{}-$\sphere$ UNet & & $0{\to}\{-2,\dots,2\}{\to}1$ & $c=4$ & $24{,}865$ & $2.52\times10^{-14}$ \\
\SSFNO{}-$\SOthree$ UNet & $8$ & $0{\to}\{-1,0,1\}{\to}1$ & $c=8$ & $19{,}762$ & $1.76\times10^{-14}$ \\
\SSFNO{}-$\SOthree$ UNet & $8$ & $1{\to}\{-1,0,1\}{\to}0$ & $c=8$ & $19{,}762$ & $1.61\times10^{-14}$ \\
\SSFNO{}-$\SOthree$ UNet & $4$ & $0{\to}\{-1,0,1\}{\to}1$ & $c=4$ & $6{,}346$ & $2.29\times10^{-14}$ \\
\bottomrule
\end{tabular}
\caption{Equivariance errors of linear models at $L=32$, measured by the $\mraee$ \eqref{eq:mraee} over the $10^4$ pairs of signal and rotation. Spins and channels are given as input, hidden and output, or as one set of spins used throughout, and $c$ is the base width of a UNet, with encoder widths $c$, $2c$, $4c$ and $4c$. $^{\ddagger}$The first block of the \SSFNO{}-$\sphere$ UNet.}
\label{tab:equiv-linear}
\end{table}

\section{Datasets}
\label{app:datasets}

We describe the three benchmarks of Sec.~\ref{sec:datasets} in turn.

\subsection{Spin spherical MNIST}
\label{app:mnist}

We project planar digits onto the sphere following \citet{cohen2018spherical} and retain their scalar harmonic coefficients up to bandlimit $L$. We also construct a Sobel variant, motivated by \citet{esteves2020spin,xu2022unified}, in which the Sobel derivatives of the projected image along the colatitude and longitude axes of the sampling grid, with their means subtracted, are combined into $X_\theta+\ii X_\varphi$, whose modulus is rescaled to $[0,1]$ for each sample with the phase kept, and the result is transformed as a spin-$1$ field before any rotation.

The harmonic variants are computed with \texttt{s2fft} \citep{price2024differentiable}, by a forward spherical harmonic transform of the scalar image at bandlimit $L$, a rotation in harmonic space by \eqref{eq:rotation-coeffs}, exact at that bandlimit, and an inverse spin spherical harmonic transform at the desired spin, which synthesises $\sw{s}{f}=\sum_{\ell=\lvert s\rvert}^{L-1}\sum_m f_{\ell m}\sw{s}{Y}_{\ell m}$. The synthesis keeps the scalar coefficients of the degrees $\ell\ge\lvert s\rvert$, without the degree-dependent factors of differential spin raising, and commutes with rotations by \eqref{eq:rotation-coeffs}. A spin-$1$ field so constructed has no monopole, whereas the scalar fields keep a small one, since the mean subtracted from each image is the unweighted mean over the grid rather than the quadrature-weighted spherical mean.

For the prediction experiments we generate $60{,}000$ samples at $L=32$ on the sampling grid of \citet{mcewen2011novel} with $33\times64$ samples, from the first $51{,}000$ images of the MNIST training set and the first $9{,}000$ of the MNIST test set, which form an $85/15$ split into training and held-out sets. Rotations of the held-out samples use Euler angles $(\alpha,\beta,\gamma)$ drawn uniformly from $[0,2\pi)^3$, which is not the Haar distribution on $\SOthree$ (that would draw $\cos\beta$ uniformly), so that the rotated test set over-represents some rotation axes. The generation code and the exact split indices will be released with \texttt{s2ai}.

\paragraph{Splits and protocol.} The classification experiments use a regenerated split of the same construction, in the image, Sobel spin-$1$, harmonic spin-$1$ and harmonic spin-$2$ variants. The first $55{,}000$ images of the MNIST training set are used for training and the remaining $5{,}000$, unrotated, for validation, and the $10{,}000$ images of the MNIST test set for testing, with Haar-distributed rotations drawn once with fixed seeds. Following \citet{cohen2018spherical}, models are trained on unrotated samples and evaluated on unrotated (NR/NR) and randomly rotated (NR/R) test samples, and all results reported in Sec.~\ref{sec:experiments:mnist} are NR/R.

\subsection{Global wind forecasting}
\label{app:data:wind}

The training set is a subset of the WeatherBench~2 $1.5^{\circ}$ archive of ERA5 reanalysis data \citep{rasp2024weatherbench2,hersbach2020era5} at the $50$~hPa pressure level, covering four six-hourly observations per day between 1980 and 2022. The observations at 00:00, 06:00 and 12:00~UTC form the input and the observation at 18:00~UTC the target, with the zonal and meridional components combined into the spin-$1$ field $u+\ii v$ (Sec.~\ref{sec:datasets}), which is the field $v_\theta+\ii v_\varphi$ of the meridian frame up to the constant factor $-\ii$. Data are converted from the equiangular latitude--longitude grid to the sampling of \citet{mcewen2011novel} at $L=128$, with $129\times256$ samples, and standardised by a real scale, after subtracting the mean of the training set. For evaluation, eight indices are drawn at random in the range $(0, 365)$, ignoring leap years, and the corresponding days, which cover all seasons, are held out from every year. The preprocessing code and the held-out indices will be released with \texttt{s2ai}.

\subsection{Weak gravitational lensing}
\label{app:massmapping}

Weak lensing is described by two fields derived from the lensing potential $\psi$, namely the spin-$0$ convergence $\kappa=\tfrac{1}{2}\bar\eth\eth\psi$, which traces the projected matter density, and the spin-$2$ shear $\gamma=\gamma_1+\ii\gamma_2=\tfrac{1}{2}\eth\eth\psi$. Since both derive from the same potential, they are related in harmonic space by the spherical Kaiser--Squires relation \citep{kaiser1993mapping,wallis2022mapping}, which in our conventions reads
\begin{equation}
  \sw{2}{\gamma}_{\ell m}=\mathcal{D}_\ell\,\kappa_{\ell m},\qquad
  \mathcal{D}_\ell=-\sqrt{\frac{(\ell+2)(\ell-1)}{\ell(\ell+1)}},\quad\ell\ge2 .
  \label{eq:lensing-forward}
\end{equation}
The forward map \eqref{eq:lensing-forward} is therefore the spin $0\to2$ layer of Theorem~\ref{thm:completeness} with $c_\ell=\mathcal{D}_\ell$, and spherical Kaiser--Squires is the spin $2\to0$ layer with $c_\ell=\mathcal{D}_\ell^{-1}$, where $\lvert\mathcal{D}_\ell^{-1}\rvert\le\sqrt{3/2}$, on the degrees $\ell\ge2$ that the shear determines. Practical considerations make the recovery of $\kappa$ ill-posed and call for priors learned from simulations.

The convergence maps are simulated with GLASS \citep{tessore2023glass} and CAMB \citep{lewis2000efficient} for a Planck 2018 cosmology \citep{planck2020cosmological}, with lognormal matter shells of $200$~Mpc from redshift $z=0$ to $1$, one Gaussian source bin at $z=0.5$, HEALPix resolution $N_{\mathrm{side}}=512$ \citep{gorski2005healpix} and $\ell_{\max}=1024$. The shear maps are forward modelled with \texttt{DarkMax}\footnote{Will be made public at \url{https://github.com/astro-informatics/darkmax}.} with the noise of a Stage IV survey \citep{laureijs2011euclid}, a galaxy number density $n_{\mathrm{gal}}=30$~arcmin$^{-2}$ and an ellipticity dispersion $\sigma_e=0.37$, with a different noise seed for each map. Maps are used at $L=256$ on the sampling grid of \citet{mcewen2011novel}, with $257\times512$ samples. The training set holds $10{,}000$ shear--convergence pairs, split $80/20$ into training and validation sets, and the evaluation set $600$ further pairs from the same pipeline.

\section{Training details and additional results}
\label{app:training}

Each model is trained once, except where stated, so that our results carry no estimate of their variation between runs.

\subsection{Spin spherical MNIST}
\label{app:training:mnist}

\paragraph{Training.} The classifiers are trained in single precision on a shared NVIDIA GB10 with Adam \citep{kingma2015adam} ($\beta_1=0.9$) at a learning rate of $10^{-3}$, multiplied by $0.2$ at $50\%$ and again at $83\%$ of the run, in batches of $32$ for $12$ epochs ($5$ or $6$ for Conv-$\SOthree$, Table~\ref{tab:mnist-classification}), minimising the categorical cross-entropy, and the reported checkpoint is the epoch of best accuracy on the non-rotated validation split. The prediction UNets are trained in double precision with Adam ($\beta_1=0.9$) at a learning rate of $5\times10^{-4}$ for up to 2000 epochs in batches of $32$, minimising the squared modulus of the difference between prediction and target, integrated with the quadrature weights of the sampling theorem, on NVIDIA GH200 and A100 GPUs with identical code. The reported prediction model is the checkpoint with the lowest loss on the rotated held-out set, on which the NR/R results are also evaluated, so that, with no separate validation set, the reported prediction errors are those of the best checkpoint on the evaluation data.

\begin{figure}[htbp]
  \centering
  \includegraphics[width=\textwidth]{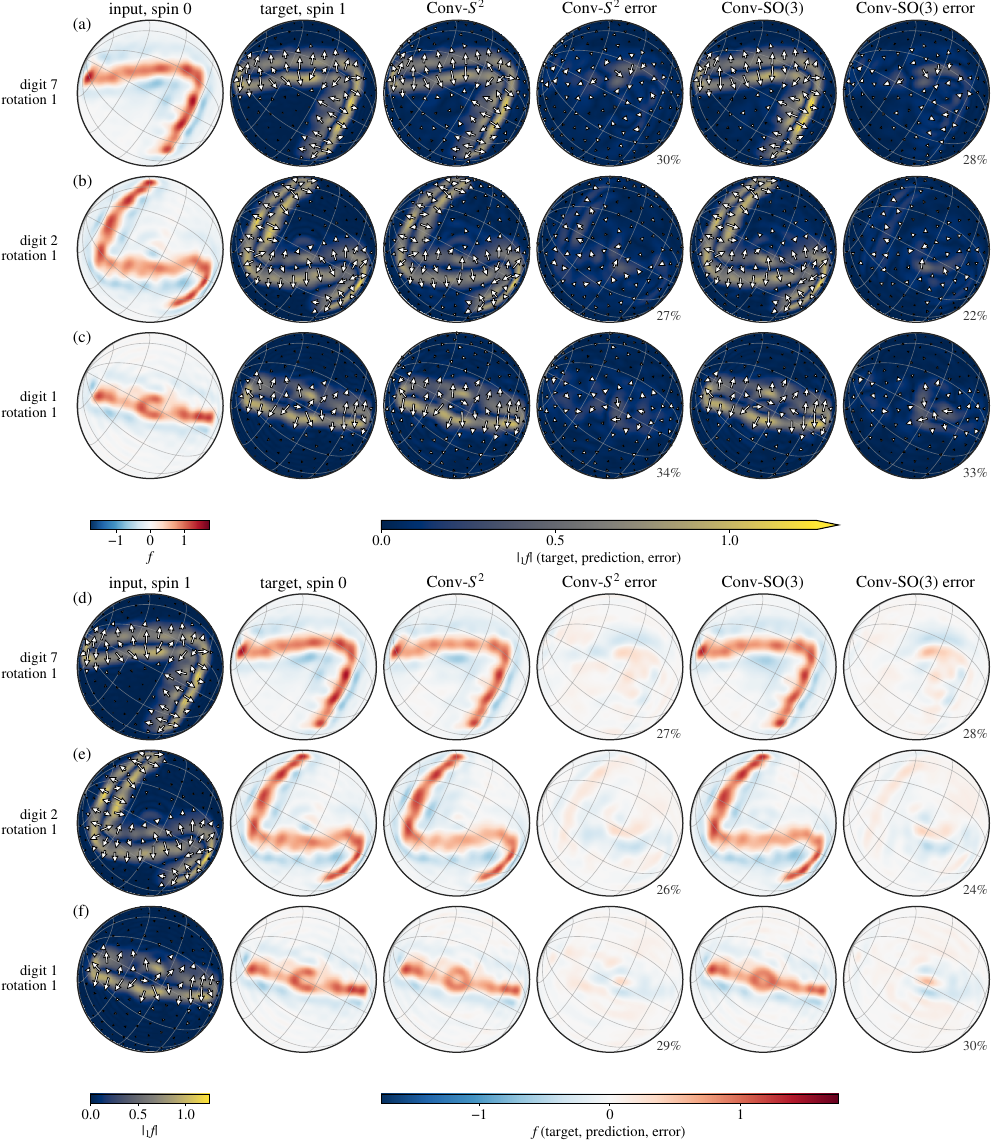}
  \caption{Prediction by the Conv-$\sphere$ and Conv-$\SOthree$ UNets at $L=32$ from the spin-$0$ image to the spin-$1$ Sobel field (a--c) and back (d--f), for the first three test digits under one rotation. Errors (prediction minus target) are on the scales of the target, spin-$1$ fields are drawn as arrows over their modulus, and percentages give the relative error $\sqrt{\mathrm{MSE}/\overline{\lvert f\rvert^2}}$.}
  \label{fig:mnist-pred-full}
\end{figure}

\paragraph{Prediction examples.} The samples of Fig.~\ref{fig:mnist-pred-full} are fixed in advance, not chosen by their errors.

\subsection{Wind forecasting}
\label{app:training:wind}

\paragraph{Training.} Both UNets of Appendix~\ref{app:architectures} are trained for 200 epochs with Adam \citep{kingma2015adam} at its default settings and a constant learning rate, minimising the mean squared error computed on the sphere, with the learning rates and batch sizes of Table~\ref{tab:wind-full}, chosen with ablation studies. The reported model of each kind is the checkpoint with the lowest loss on the held-out days.

\begin{table}[htbp]
  \caption{Wind forecasting at 50~hPa and $L=128$: configurations and all metrics on the held-out days, in physical units (MSE in m$^2$\,s$^{-2}$, RMSE and MAE in m\,s$^{-1}$, SNR and PSNR in dB). Input shapes are batch $\times$ $\theta$ $\times$ $\varphi$ $\times$ channels, and best results are in bold.}
  \label{tab:wind-full}
  \centering\small
  \setlength{\tabcolsep}{4pt}
  \begin{tabular}{lrcc cccccc}
    \toprule
    Model & Params & LR & Input shape & MSE $\downarrow$ & RMSE $\downarrow$ & MAE $\downarrow$ & SNR $\uparrow$ & PSNR $\uparrow$ & PCC $\uparrow$ \\
    \midrule
    Spin UNet     & 48{,}446 & $9 \times 10^{-4}$ & $20{\times}129{\times}256{\times}3$ & \textbf{9.52} & \textbf{3.09} & \textbf{2.67} & \textbf{14.02} & \textbf{25.06} & \textbf{0.979} \\
    Scalar UNet & 74{,}398 & $8 \times 10^{-4}$ & $50{\times}129{\times}256{\times}6$ & 17.27 & 4.16 & 3.56 & 11.43 & 22.64 & 0.969 \\
    Persistence      & -- & -- & -- & 14.55 & 3.82 & 3.31 & 12.17 & 23.19 & 0.962 \\
    \bottomrule
  \end{tabular}
\end{table}

\begin{figure}[htbp]
  \centering
  \includegraphics[width=\textwidth]{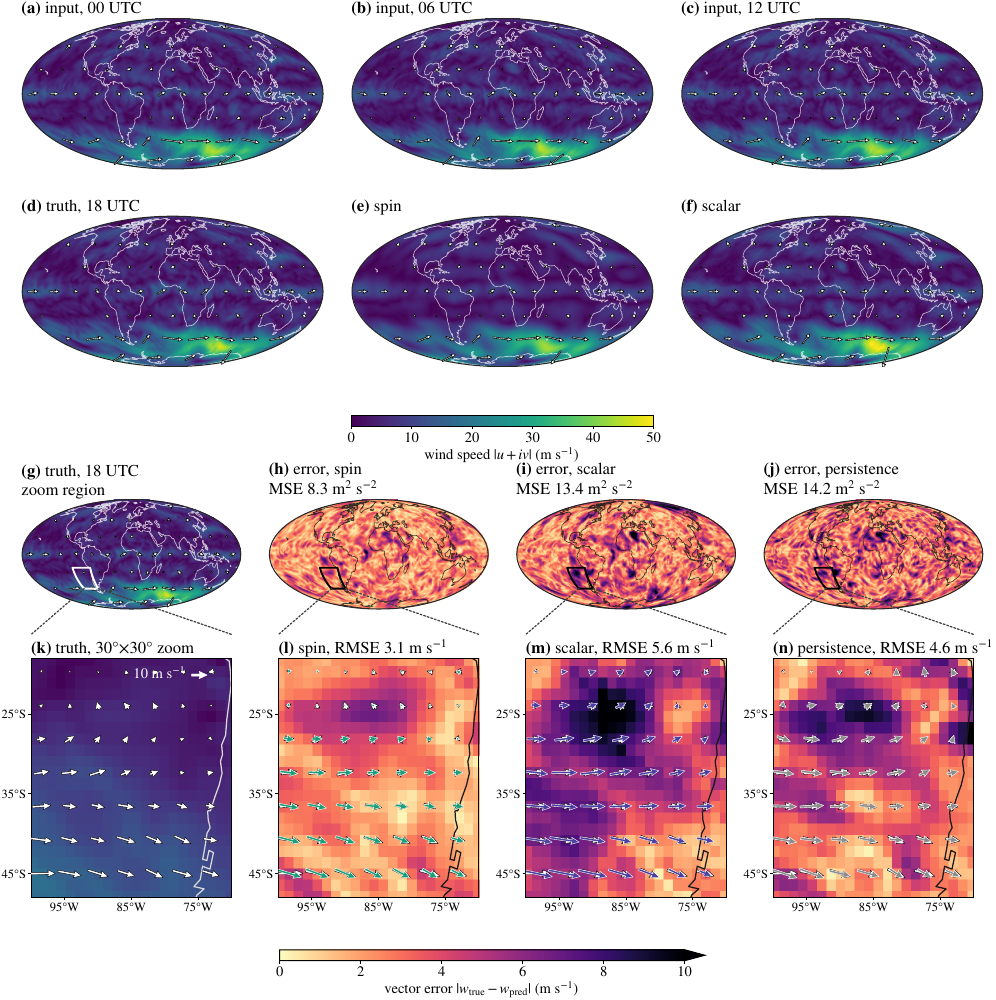}
  \caption{Wind forecast for one held-out day at $L=128$: the speed $\lvert u+\ii v\rvert$ of the inputs, the target and the forecasts (a--f), the vector error $\lvert w_{\mathrm{true}}-w_{\mathrm{pred}}\rvert$ of the spin UNet, the scalar UNet and persistence with the area-weighted MSE of this day (h--j), and a zoom into the region outlined in (g), with true (white) and predicted (coloured) vectors and the RMSE over the region (k--n).}
  \label{fig:wind-full}
\end{figure}

\subsection{Weak gravitational lensing}
\label{app:training:massmapping}

\paragraph{Training.} Both UNets of Appendix~\ref{app:architectures} are trained for 200 epochs on the $10{,}000$ shear--convergence pairs of Appendix~\ref{app:massmapping}, split $80/20$ into training and validation sets, minimising a spherically weighted mean squared error with Adam \citep{kingma2015adam}, with the inputs, batch sizes and learning rates of Table~\ref{tab:massmapping-full}. Evaluation is on the $600$ held-out maps.

\begin{table}[htbp]
  \caption{Weak lensing mass mapping at $L=256$ with shape noise: configurations and all metrics, averaged over the 600 evaluation maps. Input shapes are batch $\times$ $\theta$ $\times$ $\varphi$ $\times$ channels, and best results are in bold.}
  \label{tab:massmapping-full}
  \centering\small
  \setlength{\tabcolsep}{4pt}
  \begin{tabular}{lrcc cccccc}
    \toprule
    & & & & MSE $\downarrow$ & RMSE $\downarrow$ & MAE $\downarrow$ & SNR $\uparrow$ & PSNR $\uparrow$ & PCC $\uparrow$ \\
    Model & Params & LR & Input shape & $\times10^{6}$ & $\times10^{3}$ & $\times10^{3}$ & (dB) & (dB) & \\
    \midrule
    Spin UNet     & 80{,}702 & $5 \times 10^{-4}$ & $1{\times}257{\times}512{\times}1$ & \textbf{1.28} & \textbf{1.13} & \textbf{0.90} & \textbf{7.53} & \textbf{23.47} & \textbf{0.907} \\
    Scalar UNet & 93{,}198 & $6 \times 10^{-4}$ & $20{\times}257{\times}512{\times}2$ & 2.06 & 1.43 & 1.14 & 5.45 & 21.40 & 0.846 \\
    \bottomrule
  \end{tabular}
\end{table}

\begin{figure}[htbp]
  \centering
  \includegraphics[width=\textwidth]{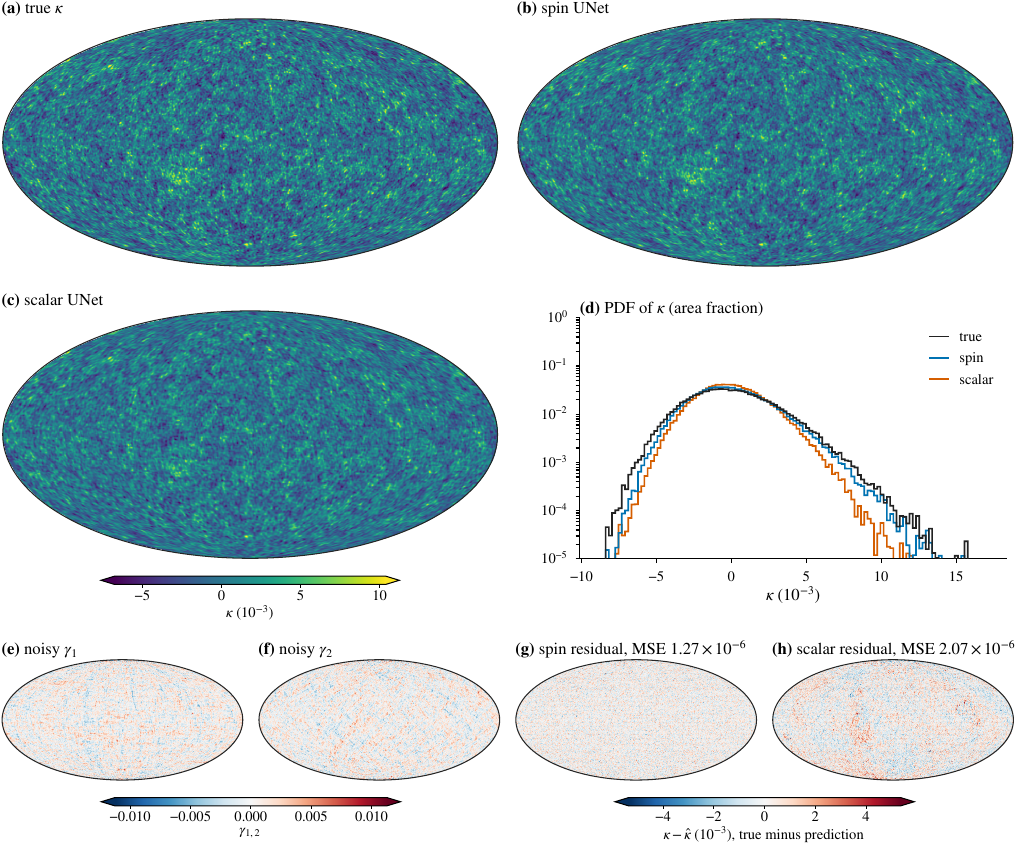}
  \caption{Mass mapping for one evaluation map at $L=256$, with the true convergence (a), its reconstructions by the spin and scalar UNets (b, c) and their one-point distributions with that of the target (d), the noisy input shear (e, f) and the residuals $\kappa-\hat\kappa$ of the two UNets with their MSE on this map (g, h). The scalar residual retains large-scale structure that the spin residual does not.}
  \label{fig:massmap-full}
\end{figure}

\paragraph{Results.} The residual of the scalar UNet retains high-amplitude, large-scale structure (Fig.~\ref{fig:massmap-full}), and the reconstruction of the spin UNet follows the tails of the one-point distribution of the target more closely (Fig.~\ref{fig:massmap-full}d).

\end{document}